\def\year{2022}\relax
\documentclass[letterpaper]{article} 

\usepackage[ruled,vlined, algo2e]{algorithm2e}
\usepackage[table]{xcolor}

\usepackage{stackengine}

\SetAlFnt{\footnotesize}

\SetCommentSty{xCommentSty}

\SetNoFillComment

\SetNlSty{mynlfont}{}{} 

\LinesNumbered

\SetSideCommentRight

\DontPrintSemicolon

\RestyleAlgo{algoruled}

\SetKwProg{Fn}{def}{}{}{}

\SetKwInOut{Input}{In}
\SetKwInOut{Output}{Out}

\makeatletter

\newcommand{\checknextarg}{\@ifnextchar\bgroup{\gobblenextarg}{}}
\newcommand{\gobblenextarg}[1]{ $\leftarrow$ $#1$\@ifnextchar\bgroup{\gobblenextarg}{}}

\usepackage{amsmath,amssymb,mathrsfs,mathtools,lipsum,amsthm}
\theoremstyle{definition}

\usepackage{cuted}
\usepackage{color}
\usepackage{setspace}

\newcommand{\brcomment}[1]{}

\newcommand{\ie}{{\it i.e.,}}
\newcommand{\eg}{{\it e.g.,}}

\newcounter{magicrownumbers}

\newcommand{\RNum}[1]{\uppercase\expandafter{\romannumeral #1\relax}}
\newcommand{\Srel}[1]{\mathrel{{}^{#1}\kern-\scriptspace}}

\usepackage{array}
\newcolumntype{L}[1]{>{\raggedright\let\newline\\\arraybackslash\hspace{0pt}}m{#1}}
\newcolumntype{C}[1]{>{\centering\let\newline\\\arraybackslash\hspace{0pt}}m{#1}}
\newcolumntype{R}[1]{>{\raggedleft\let\newline\\\arraybackslash\hspace{0pt}}m{#1}}

\usepackage{enumitem}
\usepackage{textcomp}
\usepackage{booktabs}

\usepackage{graphicx}
\graphicspath{ {images/} }

\usepackage{caption}
\usepackage{subcaption}

\usepackage{wrapfig}

\usepackage{upgreek}
\usepackage{bm} 

\usepackage{cuted}
\usepackage{multirow}
\usepackage{svg}
\usepackage{aaai22}  
\usepackage{times}  
\usepackage{helvet}  
\usepackage{courier}  
\usepackage[hyphens]{url}  
\usepackage{graphicx} 
\usepackage{natbib}  
\usepackage{caption} 
\DeclareCaptionStyle{ruled}{labelfont=normalfont,labelsep=colon,strut=off} 
\usepackage{algorithm}
\usepackage{algorithmic}

\usepackage{newfloat}
\usepackage{listings}
\floatstyle{ruled}
\newfloat{listing}{tb}{lst}{}
\floatname{listing}{Listing}
\title{CTIN: Robust Contextual Transformer Network for Inertial Navigation}

\author{
    Bingbing Rao\textsuperscript{\rm 1},
    Ehsan Kazemi\textsuperscript{\rm 1, 2},
    Yifan Ding\textsuperscript{\rm 1},
    Devu M Shila\textsuperscript{\rm 2},
    Frank M. Tucker\textsuperscript{\rm 3},
    Liqiang Wang\textsuperscript{\rm 1}
}
\affiliations{

    \textsuperscript{\rm 1}Department of Computer Science, University of Central Florida, Orlando, FL, USA\\
    \textsuperscript{\rm 2} Unknot.id Inc., Orlando, FL, USA \\
    \textsuperscript{\rm 3} U.S. Army CCDC SC, Orlando, FL, USA
}

\begin{document}

\maketitle

\thispagestyle{plain}
\pagestyle{plain}

\begin{abstract}

Recently, data-driven inertial navigation approaches have demonstrated their capability of using well-trained neural networks to obtain accurate position estimates from inertial measurement units (IMUs) measurements. In this paper, we propose a novel robust {\bf C}ontextual {\bf T}ransformer-based network for {\bf I}nertial {\bf N}avigation (CTIN) to accurately predict velocity and trajectory. To this end, we first design a ResNet-based encoder enhanced by local and global multi-head self-attention to capture spatial contextual information from IMU measurements. Then we fuse these spatial representations with temporal knowledge by leveraging multi-head attention in the Transformer decoder. Finally, multi-task learning with uncertainty reduction is leveraged to improve learning efficiency and prediction accuracy of velocity and trajectory. Through extensive experiments over a wide range of inertial datasets (\eg\; RIDI, OxIOD, RoNIN, IDOL, and our own), CTIN is very robust and outperforms state-of-the-art models.


\end{abstract}

\section{Introduction}

Inertial navigation is a never-ending endeavor to estimate the states (\ie \; position and orientation) of a moving subject (\eg\; pedestrian) by using only IMUs attached to it. An IMU sensor, often a combination of accelerometers and gyroscopes, plays a significant role in a wide range of applications from mobile devices to autonomous systems because of its superior energy efficiency, mobility, and flexibility \cite{lymberopoulos2015realistic}. 
Nevertheless, the conventional Newtonian-based inertial navigation methods reveal not only poor performance, but also require unrealistic constraints that are incompatible with everyday usage scenarios. For example, strap-down inertial navigation systems (SINS) may obtain erroneous sensor positions by performing double integration of IMU measurements, duo to exponential error propagation through integration \cite{titterton2004strapdown}. Step-based pedestrian dead reckoning (PDR) approaches can reduce this accumulated error by leveraging the prior knowledge of human walking motion to predict trajectories \cite{tian2015enhanced}. However, an IMU must be attached to a foot in the zero-velocity update \cite{foxlin2005pedestrian} or a subject must walk forward so that the motion direction is constant in the body frame \cite{brajdic2013walk}. In addition, inertial sensors are often combined with additional sensors and models using Extended Kalman Filter \cite{bloesch2015robust} to provide more accurate estimations, where the typical sensors include WiFi \cite{ahmetovic2016navcog}, Bluetooth \cite{li2017indoor}, LiDAR \cite{zhang2014loam}, or camera sensors \cite{leutenegger2015keyframe}. Nonetheless, these combinations with additional sensors are posing new challenges about instrument installations, energy efficiency, and data privacy. For instance, Visual-Inertial Odometry (VIO) substantially depends on environmental factors such as lighting conditions, signal quality, blurring effects \cite{usenko2016direct}. 



Recently, a growing number of data-driven approaches such as IONet \cite{chen2018ionet}, RoNIN \cite{herath2020ronin}, and IDOL \cite{sun2021idol} have demonstrated their capability of using well-trained neural networks to obtain accurate estimates from IMU measurements with competitive performance over the aforementioned methods. However, grand challenges still exist when applying neural network techniques to IMU measurements: 1) most existing data-driven approaches leverage sequence-based models (\eg\; LSTM \cite{LSTM}) to learn temporal correlations but fail to capture spatial relationships between multivariate time-series. 2) There is few research work to explore rich {\it contextual} information among IMU measurements in dimensions of spatial and temporal for inertial feature representation. 3) Usually, uncertainties of IMU measurements and model output are assumed to be a fixed covariance matrix in these pure and black-box neural inertial models, which brings significant inaccuracy and much less robustness because they can fluctuate dramatically and unexpectedly in nature. 





In response to the observations and concerns raised above, a novel robust contextual Transformer network is proposed to regress velocity and predict trajectory from IMU measurements. Particularly, CTIN extends the ideas of ResNet-18 \cite{he2016deep} and Transformer \cite{vaswani2017attention} to exploit spatial and longer temporal information among IMU observations and then uses the attention technique to fuse this information for inertial navigation. The major contributions of this paper are summarized as follows:
\begin{itemize}
    \item Extending ResNet-18 with attention mechanisms is to explore and encode spatial information of IMU samples.
    
    \item A novel self-attention mechanism is proposed to extract contextual features of IMU measurements.
    
    \item Multi-Task learning using novel loss functions is to improve learning efficiency and reduce models' uncertainty. 
    
    \item Comprehensive qualitative and quantitative comparisons with the existing baselines indicate that CTIN outperforms state-of-the-art models.
      
    
    \item A new IMU dataset with ground-truth trajectories under natural human motions is provided for future reference.
    
    \item To the best of our knowledge, CTIN is the first Transformer-based model for inertial navigation.
        
\end{itemize}

The rest of this paper is organized as follows. Section \ref{sec: background} gives background about inertial navigation and attention mechanism. Section \ref{sec: model} reviews architecture and workflow of CTIN. The evaluation and related work are discussed in Section \ref{sec: evaluation} and Section \ref{sec: related_work}. Section \ref{sec: conclusion} introduces conclusion and future work.

\section{Background}
\label{sec: background}

\subsection{IMU models} 

Technically, 3D angular velocity ($\omega$) and 3D acceleration ($\alpha$) provided by IMUs are subjected to bias and noise based on some sensor properties, as shown in Equation \ref{eq:1} \& \ref{eq:3}: 
\begin{small}\begin{align}
  \omega_{t}&=r^{\omega}_{t}+ b^{\omega}_{t} + n^{\omega}_{t} \label{eq:1}\\
  \alpha_{t}&=r^{\alpha}_{t}+ b^{\alpha}_{t} + n^{\alpha}_{t} \label{eq:3}
\end{align}\end{small}where $r^{\omega}_{t}$ and $r^{\alpha}_{t}$ are real sensor values measured by the gyroscope and accelerometer at timestamp $t$, respectively; $b^{\omega}_t$ and $b^{\alpha}_t$ are time-varying bias; $n^{\omega}_t$ and $n^{\alpha}_t$ are noise values, which usually follow a zero-mean gaussian distribution.





\subsection{Inertial Tracking}

According to Newtonian mechanics \cite{kok2017using}, states (\ie \; position and orientation) of a moving subject (\eg\; pedestrian) can be estimated from a history of IMU measurements, as shown in Equation \ref{eq:kinematic_model}:
\begin{small}\begin{subequations}
    \label{eq:kinematic_model}
	\begin{align}
	    \label{subeq:ori}
		R_b^n(t) = R_b^n(t - 1) \otimes \Omega(t)\\
		\label{subeq:rel_ori}
	    \Omega(t) = exp(\frac{dt}{2}\omega(t-1))\\
    	\label{subeq:vel}
    	v^n(t) = v^n(t - 1) + \Delta(t) \\
    	\label{subeq:rel_vel}
    	\Delta(t) = (R_b^n(t-1) \odot \alpha(t-1) - g^n)dt \\
    	\label{subeq:position}
    	P^n(t) = P^n(t-1) + v^n(t-1)dt
	\end{align}
\end{subequations}\end{small}Here, the orientation $R_b^n(t)$ at timestamp $t$ is updated with a relative orientation ($\Omega(t)$) between two discrete instants $t$ and $t-1$ according to Equation \ref{subeq:ori} \& \ref{subeq:rel_ori}, where $\omega(t-1)$ measures proper angular velocity of an object at timestamp $(t-1)$ in the body frame (denoted by $b$) with respect to the navigation frame (denoted by $n$). 
$R_b^n$ can be used to rotate a measurement $x \in [\omega, \alpha]$ from the body frame $b$ to the navigation frame $n$, which is denoted by an expression $R_b^n \odot x$  = $R_b^n \otimes x \otimes (R_b^{n})^T$ where $\otimes$ is a hamilton product between two quaternions. The navigation frame in our case is defined such that Z axis is aligned with earth's gravity $g^n$ and the other two axes are determined according to the initial orientation of the body frame. In Equation \ref{subeq:vel} \& \ref{subeq:rel_vel}, velocity vector $v^n(t)$ is updated with its temporal difference $\Delta(t)$, which is obtained by rotating $\alpha(t-1)$ to the navigation frame using $R_b^w(t-1)$ and discarding the contribution of gravity forces $g^n$. Finally, positions $P^n(t)$ are obtained by integrating velocity in Equation \ref{subeq:position}. Therefore, given current IMU measurements (\ie\; $\alpha$, $\omega$), the new system states (\ie\; $P^n$, $v^n$ and $R_b^n$) can be obtained from the previous states using a function of $f$ in Equation \ref{eq:sts_state}, where $f$ represents transformations in Equation \ref{eq:kinematic_model}.
\begin{small}\begin{equation}
    \label{eq:sts_state}
    [P^n, v^n, R_b^n]_t = f([P^n, v^n, R_b^n]_{t-1}, [\alpha, \omega]_t),
\end{equation}\end{small}{\bf Drawback and Solution:} However, using IMUs for localization results in significant drift due to that the bias and noise intrinsic to the gyroscope and accelerometer sensing can explode quickly in the double integration process. Using pure data-driven models with IMU measurements for Inertial Navigation has shown promising results in pedestrian dead-reckoning systems. To tackle the problems of error propagation in Equation \ref{eq:sts_state}, we break the cycle of continuous integration and segment inertial measurements into independent windows, then leverage a sequence-to-sequence neural network architecture \cite{sutskever2014sequence, bahdanau2014neural, wu2016google, vaswani2017attention} to predict velocities and positions from an input window $m$ of IMU measurements, as shown in Equation \ref{eq:sts_state_w}.
\begin{small}\begin{equation}
    \label{eq:sts_state_w}
    [P^n, v^n]^{1:m} = \mathcal{F}_{\theta}(P^n_0, v^n_0, [R_b^n, \alpha, \omega]^{1:m}),
\end{equation}\end{small}where $\mathcal{F}_{\theta}$ represents a latent neural system that learns the transformation from IMU samples to predict positions and velocities, where $P^n_0$, $v^n_0$ are initial states.

\begin{figure*}
    \centering
    \includegraphics[scale=0.75]{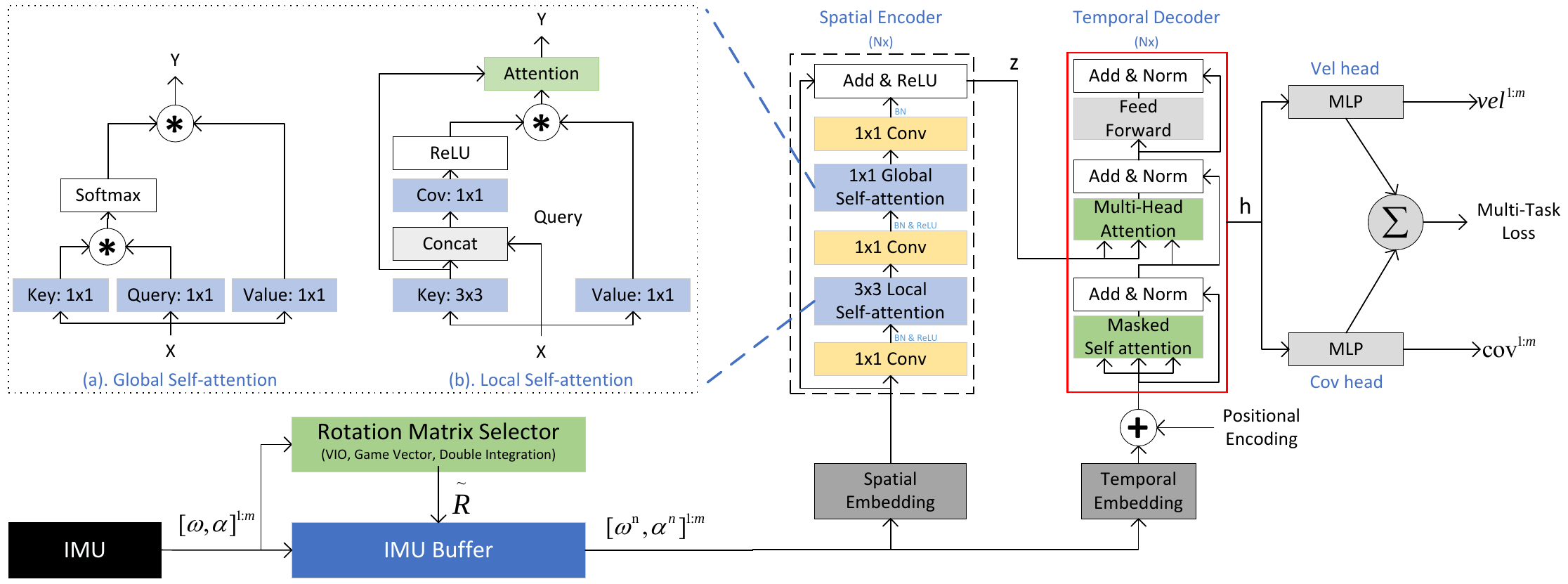}
    \vspace{-2mm}
    \caption{Overall workflow of the proposed contextual transformer model for inertial navigation. 
    }
    \label{fig:overall_arch}
    \vspace{-4mm}
\end{figure*}

\subsection{Attention Mechanism}
{\it Attention} can be considered as a query procedure that maps a query $Q$ for a set of key-value pairs ($K, V$) to an output \cite{vaswani2017attention, han2020survey}, which is denoted by $ATT(Q, K, V) = \gamma(Q, K) \times V$. Typically, the output is computed as a sum of weighted values ($V$), where the weights $\gamma(Q, K)$ are computed according to a compatibility function of $Q$ and $K$. There are two kinds of $\gamma$ used in this paper \cite{bahdanau2014neural, wang2018non}: (1) we perform a {\it dot product} between $Q$ and $K$, divides each resulting element by $\sqrt{d}$, and applies a {\it softmax} function to obtain the weights: $\gamma(Q, K) = softmax(\frac{QK^T}{\sqrt{d}})$ where $d$ is the dimension size of vectors $Q$, $K$ and $V$. (2) Inspired by Relation Networks \cite{santoro2017simple}, we investigate a form of {\it concatenation}: $\gamma(Q, K) = ReLU(W_{\gamma}[Q, K])$, where $[\cdot,\cdot]$ denotes concatenation and $W_{\gamma}$ is a weight vector that projects the concatenated vector to a scalar. {\it Self-attention} networks compute a representation of an input sequence by applying attention to each pair of tokens from the sequence, regardless of their distance \cite{vaswani2017attention}. Technically, given IMU samples $X \in \mathbb{R}^{m \times d}$, we can perform the following transformation on $X$ directly to obtain $Q$, $K$ and $V$: $Q, K, V = XW_Q, XW_K, XW_V$, where $\{W_Q, W_K, W_V\} \in R^{d \times d}$ are trainable parameters. Usually, these intermediate vectors are split into different representation subspaces at different positions (\ie\;$h=8, d_k = \frac{d}{h}$), \eg\;$K = [K^1, \ldots, K^h]$ with $K^i \in \mathbb{R}^{m\times d_k}$. For a subspace, the attention output is calculated by $head_i = ATT(Q^i, K^i, V^i)$. The final output representation is the concatenation of outputs generated by multiple attention heads: $MultiHead(Q, K, V) = [head_i, \ldots, head_h]$.







\section{Our Approach}
\label{sec: model}




\subsection{System Overall}
\label{subsec:problem_statement}






The Attention-based architecture for inertial navigation is shown in Figure \ref{fig:overall_arch} and its workflow is depicted as follows:

{\bf Data Preparation.} Initially, an IMU sample is the concatenation of data from gyroscope and accelerometer. To exploit temporal characteristics of IMU samples, we leverage a sliding window with size $m$ to prepare datasets at timestamp $t$, denoted by $X^{1:m}_{t} = [x_{t - m + 1}, \ldots, x_t]$. Similarly, we adopt this rolling mechanism with the same window size to build the ground truth of velocities: $gt_{vel}^{1:m}$. Usually, IMU samples in each window are rotated from the body frame (\ie\;$\omega^b, \alpha^b$) to the navigation frame (\ie\;$\omega^n, \alpha^n$) using provided orientations. {\it Rotation Matrix Selector} is designed to select sources of orientation for training and testing automatically. Typically, we use the device orientation estimated from IMU for testing.

{\bf Embedding.} We need to compute feature representations for IMU samples before feeding them into encoder and decoder. {\it Spatial Embedding} uses a 1D convolutional neural network followed by batch normalization and linear layers to learn spatial representations; {\it Temporal Embedding} adopts a 1-layer bidirectional LSTM model to exploit temporal information, and then adds positional encoding provided by a trainable neural network.

{\bf Spatial Encoder.} The encoder comprises a stack of $N$ identical layers, which maps an input sequence of $X^{1:m}_{t}$ to a sequence of continuous representations $z = (z_1, \dots, z_m)$. To capture spatial knowledge of IMU samples at each timestamp, we strengthen the functionality of the core bottleneck block in ResNet-18 \cite{he2016deep} by replacing spatial convolution with a local self-attention layer and inserting a global self-attention module before the last $1\times1$ downsampling convolution (cf. in Section \ref{subsec: att}). All other structures, including the number of layers and spatial downsampling, are preserved. The modified bottleneck layer is repeated multiple times to form {\it Spatial Encoder}, with the output of one block being the input of the next one. 

{\bf Temporal Decoder.} The decoder also comprises a stack of $N$ identical layers. Within each layer, we first perform a masked self-attention sub-layer to extract dependencies in the temporal dimension. The masking emphasizes a fact that the output at timestamp $t$ can depend only on IMU samples at timestamp less than $t$. Next, we conduct a multi-head attention sub-layer over the output of the encoder stack to fuse spatial and temporal information into a single vector representation and then pass through a position-wise fully connected feed-forward sub-layer. We also employ residual connections around each of the sub-layers, followed by layer normalization.

{\bf Velocity and Covariance.} Finally, two MLP-based branch heads regress 2D velocity ($vel^{1:m}_{t}$) and the corresponding covariance matrix ($cov^{1:m}_{t}$) using the input of $h$, respectively. Position can be obtained by the integration of velocity. The model of the covariance, denoted by $\boldsymbol{\Sigma}: x \rightarrow \mathbb{R}^{2\times2}$ where $x$ is a system state, can describe the distribution difference between ground-truth velocity and the corresponding predictions of them during training. Given that, the probability of a velocity $y_v$ considering current system state $x$ can be approximated by a multivariate Gaussian distribution \cite{russell2021multivariate}:
\begin{small}\begin{equation}
    \begin{split}
        p_c(y_v|x) &= \frac{1}{\sqrt{(2\pi)^2|\boldsymbol{\Sigma}(x)|}} \; \times \\          & \;\;\;\;exp(-\frac{1}{2}(y_v-\mathcal{F}_\theta(x))^T\boldsymbol{\Sigma(x)}^{-1}(y_v-\mathcal{F}_\theta(x)))
    \end{split}
     \label{eq:cov_matrix}
\end{equation}\end{small}It is worthwhile to mention that we also leverage multi-task learning with uncertainty reduction to accomplish the desired performance (See details in Section \ref{subsec: multi_task}).

\begin{table*}
\centering
\resizebox{1\linewidth}{!}{%
\begin{tabular}{lccccccll}
\toprule
\multirow{2}{*}{Dataset} & \multirow{2}{*}{Year} & \multirow{2}{*}{\shortstack{IMU \\ Carrier}} & \multirow{2}{*}{\shortstack{Sample  \\ Frequency}} & \multirow{2}{*}{\shortstack{$\textrm{N}^{\underline{o}}$ of \\ Subjects}}  &  \multirow{2}{*}{\shortstack{$\textrm{N}^{\underline{o}}$ of \\ Sequences}} & \multirow{2}{*}{\shortstack{Ground  \\ Truth}} & \multirow{2}{*}{\shortstack{Motion \\ Context}}  & \multirow{2}{*}{\shortstack{Source}}\\ 
                         &  &  &  &    & &  & &\\
\midrule

\multirow{2}{*}{RIDI}&	\multirow{2}{*}{2017}&	\multirow{2}{*}{Lenovo Phab2 Pro}	&\multirow{2}{*}{200 Hz}	&\multirow{2}{*}{10} &\multirow{2}{*}{98}	&\multirow{2}{*}{Google Tango phone}	&\multirow{2}{*}{\shortstack[l]{Four attachments:  leg pocket, \\ bag, hand, body}}  & Public \cite{yan2018ridi}\\ 
& \\\hline

\multirow{2}{*}{OxIOD}&	\multirow{2}{*}{2018}&	\multirow{2}{*}{\shortstack[c]{iPhone 5/6, 7 Plus, \\ Nexus 5}}	&\multirow{2}{*}{100 Hz}	&\multirow{2}{*}{5}	&\multirow{2}{*}{158}	&\multirow{2}{*}{Vicon}	&\multirow{2}{*}{\shortstack[l]{Four attachments:  handheld, pocket, \\ handbag, trolley  }} & Public \cite{chen2018oxiod}\\ 
& \\\hline

\multirow{1}{*}{RoNIN}&	\multirow{1}{*}{2019}&	\multirow{1}{*}{\shortstack[c]{Galaxy S9, Pixel 2 XL}}	&\multirow{1}{*}{200 Hz}	&\multirow{1}{*}{100}	& \multirow{1}{*}{276}&	\multirow{1}{*}{Asus Zenfone AR}	& \multirow{1}{*}{Attaching devices naturally}  & Public \cite{herath2020ronin}\\ 
\hline

\multirow{1}{*}{IDOL}&	\multirow{1}{*}{2020}&	\multirow{1}{*}{iPhone 8}	&\multirow{1}{*}{100 Hz}	&\multirow{1}{*}{15}	&\multirow{1}{*}{84}	&\multirow{1}{*}{Kaarta Stencil}	& \multirow{1}{*}{Attaching devices naturally}  & Public \cite{sun2021idol}\\ 
\hline

\multirow{1}{*}{CTIN}&	\multirow{1}{*}{2021}	& \multirow{1}{*}{Samsung Note, Galaxy}	&\multirow{1}{*}{200 Hz}	&\multirow{1}{*}{5}	&\multirow{1}{*}{100}	&\multirow{1}{*}{Google ARCore}	&  \multirow{1}{*}{Attaching devices naturally} & Collected by our own and will be released soon\\ 
\bottomrule
\end{tabular}
}%
\vspace{-2mm}
\caption{Description of public datasets used for evaluation of navigation models.}
\label{tab: data_desc}
\vspace{-6mm}
\end{table*}

\subsection{Attention In Inertial Navigation}
\label{subsec: att}


In this paper, the encoder and decoder rely entirely on attention mechanism with different settings for embedding matrix $\{W_Q, W_K, W_V\}$ and $\gamma$ to explore spatial and temporal knowledge from IMU samples.

{\bf Global self-attention in Encoder.} It triggers the feature interactions across different spatial locations, as shown in Figure \ref{fig:overall_arch}(a). Technically, we first transform $X$ into $Q$, $K$, and $V$ using three separated 1D $1\times1 $ convolutions, respectively. After that, we obtain the global attention matrix (\ie\;$\gamma(Q,K)$) between $K$ and $Q$ using a {\it Dot Product} version of $\gamma$. Finally, the final output $Y$ is computed by $\gamma(Q, K) \times V$. In addition, we also adopt multi-head attention to jointly summarize information from different sub-space representations at different spatial positions. 

{\bf Local self-attention in Encoder.} Although performing a global self-attention over the whole feature map can achieve competitive performance, it not only scales poorly but also misses contextual information among neighbor keys. Because it treats queries and keys as a group of isolated pairs and learns their pairwise relations independently without exploring the rich contexts between them. To alleviate this issue, a body of research work  \cite{hu2019local, ramachandran2019stand, zhao2020exploring, li2021contextual, yao2022adcnn} employs self-attention within the local region (\ie\; $3\times3$ grid) to boost self-attention learning efficiently, and strengthen the representative capacity of the output aggregated feature map. In this paper, we follow up this track and design a novel local self-attention for inertial navigation, as shown in Figure \ref{fig:overall_arch}(b). In particular, we first employ $3\times3$ group convolution over all the neighbor keys within a grid of $3\times3$ to extract local contextual representations for each key, denoted by $C_1 = XW_{K, 3\times3}$. After that, the attention matrix (\ie\;$\gamma(Q, C_1)$) is achieved through a {\it concatenation} version of $\gamma$ in which $W_{\gamma}$ is a $1\times1$ convolution and $Q$ is defined as $X$. Next, we calculate the attended feature map $C_2$ by $\gamma(Q, C_1) \times V$, which captures the global contextual interactions among all IMU samples. The final output $Y$ is fused by an attention mechanism between local context $C_1$ and global context $C_2$.

{\bf Multi-head attention in Decoder.} We inherit settings from vanilla Transformer Decoder for attention mechanisms \cite{vaswani2017attention}. In other words, we take three separated linear layers to generate $Q$, $K$ and $V$ from $X$, respectively, and leverage a pairwise function of {\it Dot product} to calculate attention matrix (\ie\;$\gamma(Q, K)$). Finally, the final output $Y$ is computed by $\gamma(Q, K) \times V$.


\subsection{Jointly Learning Velocity and Covariance}
\label{subsec: multi_task}

We leverage multi-task learning with uncertainty reduction to improve learning efficiency and prediction accuracy of the two regression tasks: prediction of 2D velocity and its covariance. Inspired by \cite{kendall2018multi, liu2020tlio, yao2021active, yang2021aevrnet}, we derive a multi-task loss function by maximizing the Gaussian likelihood with uncertainty \cite{kendall2017uncertainties}. First, we define our likelihood as a Gaussian with mean given by the model output as $p_u(y|\mathcal{F}_\theta(x)) = \mathcal{N}(\mathcal{F}_\theta(x), \delta^2)$, where $\delta$ is an observation noise scalar. Next, we derive the model's minimization objective as a Negative Log-Likelihood (NLL) of two model outputs $y_v$ (velocity) and $y_c$ (covariance): $\mathcal{L}(\mathcal{F}_\theta, \delta_v, \delta_c)$ 
\begin{small}
\begin{equation}
\vspace{-2mm}
\begin{split}
&= -\log(p_u(y_{v}, y_{c}|\mathcal{F}_\theta(x))) \\
&= -\log(p_u(y_{v}|\mathcal{F}_\theta(x)) \times p_u(y_{c}|\mathcal{F}_\theta(x))) \\
& = -(\log(p_u(y_{v}|\mathcal{F}_\theta(x))) + \log(p_u(y_{c}|\mathcal{F}_\theta(x))) \\
& = -(\log(\mathcal{N}(y_v; \mathcal{F}_\theta(x), \delta^2_v)) + \log(\mathcal{N}(y_c; \mathcal{F}_\theta(x), \delta^2_c))) \\
&\propto \underbrace{\frac{1}{2\delta_v^2} \parallel y_v - \mathcal{F}_\theta(x) \parallel^2 + \log\delta_v}_\textrm{Velocity} + \underbrace{\frac{1}{2\delta_c^2} \parallel y_c - \mathcal{F}_\theta(x) \parallel^2 + \log\delta_c}_\textrm{Covariance}\\
&= \frac{1}{2\delta_v^2} \mathcal{L}_v + \frac{1}{2\delta_c^2} \mathcal{L}_c + \log\delta_v\delta_c
\end{split}
\vspace{-2mm}
\end{equation}
\end{small}where $\delta_v$ and $\delta_c$ are observation noises for velocity and covariance, respectively. Their loss functions are denoted by $\mathcal{L}_v$ and $\mathcal{L}_c$, and depicted as follows:

\noindent{\bf Integral Velocity Loss (IVL, $\mathcal{L}_v$).} Instead of performing mean square error (MSE) between predicted velocity ($\hat{v}$) and the ground-truth value ($v$), we first integrate predicted positions from $\hat{v}$ (cf. Equation \ref{subeq:position}), and then define a L2 norm against the ground-truth positional difference within same segment of IMU samples, denoted by $\mathcal{L}^p_v$. In addition, we calculate cumulative error between $\hat{v}$ and $v$, denoted by $\mathcal{L}^{e}_v$. Finally, $\mathcal{L}_v$ is defined as $\mathcal{L}^p_v + \mathcal{L}^{e}_v$. 

\noindent{\bf Covariance NLL Loss (CNL, $\mathcal{L}_c$).} According to the covariance matrix in Equation \ref{eq:cov_matrix}, We define the Maximum Likelihood loss as the NLL of the velocity with consideration of its corresponding covariance $\boldsymbol{\Sigma}$:
\begin{small}\begin{equation}
        \begin{split}
            \mathcal{L}_c &= -\log(p_c(y_v|x)) \\
            &=\frac{1}{2}(y_v-f(x))^T\boldsymbol{\Sigma(x)}^{-1}(y_v-f(x)) + \frac{1}{2}\ln|\boldsymbol{\Sigma(x)}| \\
            &=\frac{1}{2}\parallel y_v - f(x) \parallel^2_{\boldsymbol{\Sigma(x)}} + \frac{1}{2}\ln|\boldsymbol{\Sigma(x)}|
        \end{split}
\end{equation}\end{small}There is a rich body of research work to propose various covariance parametrizations for neural network uncertainty estimation \cite{ liu2020tlio, russell2021multivariate}. In this study, we simply define the variances along the diagonal, which are parametrized by two coefficients of a velocity.

\begin{table*}
\centering
\resizebox{0.85\linewidth}{!}{%
  \begin{tabular}{c|c|c|c|c|c|c|c|c|c||c|c|c}
    \toprule
    \multirow{3}{*}{Dataset} & \multirow{3}{*}{\shortstack{Test \\Subject}} & \multirow{3}{*}{Metric} & \multicolumn{7}{c||}{Performance (meter)} & \multicolumn{3}{c}{Perf. Improvement} \\ \cline{4-13}
      & & & \multirow{2}{*}{SINS} & \multirow{2}{*}{PDR}  & \multirow{2}{*}{RIDI} & \multicolumn{3}{c|}{RoNIN} & \multirow{2}{*}{CTIN} & \multicolumn{3}{c}{CTIN improvement over RoNIN}  \\ \cline{7-9} \cline{11-13}
      & & &  &   &  & R-LSTM & R-ResNet & R-TCN &  &  R-LSTM & R-ResNet & R-TCN \\\midrule

    \multirow{6}{*}{RIDI} & \multirow{3}{*}{Seen} & ATE & 6.34  &22.76  &8.18  &2.55  &2.33  &3.25  &{\bf 1.39}  &45.36\%  &40.10\%  &57.13\% \\\cline{3-13} 
    & & T-RTE & 8.13  &24.89  &9.34  &2.34  &2.36  &2.64  &{\bf 1.99}  &15.00\%  &15.78\%  &24.80\%\\\cline{3-13}
    & & D-RTE & 0.52  &1.39  &0.97  &0.16  &0.16  &0.17  &{\bf 0.11}  &32.47\%  &32.26\%  &35.91\% \\\cline{2-13}
    & \multirow{3}{*}{Unseen} & ATE & 4.62  &20.56  &8.18  &2.78  &1.97  &2.06  &{\bf 1.86}  &33.07\%  &5.40\%  &9.68\% \\\cline{3-13}
    & & T-RTE & 4.58  &31.17  &10.51  &2.95  &2.47  &{\bf 2.43}  &2.49  &15.66\%  &-0.70\%  &-2.36\%\\\cline{3-13}
    & & D-RTE & 0.36  &1.19  &1.09  &0.15  &0.14  &0.14  &{\bf 0.11}  &28.00\%  &21.22\%  &22.72\% \\ \midrule
        
    \multirow{6}{*}{OxIOD} & \multirow{3}{*}{Seen} & ATE & 15.36  &9.78  &3.78  &3.87  &2.40  &3.33  &{\bf 2.32}  &40.10\%  &3.52\%  &30.27\% \\\cline{3-13} 
    & & T-RTE & 11.02  &8.51  &3.99  &1.56  &1.83  &1.49  &{\bf 0.62}  &60.40\%  &66.27\%  &58.67\%\\\cline{3-13}
    & & D-RTE & 0.96  &1.16  &2.30  &0.20  &0.56  &0.19  &{\bf 0.07}  &61.94\%  &86.67\%  &61.21\% \\\cline{2-13}
    & \multirow{3}{*}{Unseen} & ATE & 13.90  &17.72  &7.16  &5.22  &3.51  &6.16  &{\bf 3.34}  &35.90\%  &4.61\%  &45.69\% \\\cline{3-13}
    & & T-RTE & 10.51  &17.21  &7.65  &2.65  &2.51  &2.61  &{\bf 1.33}  &50.00\%  &47.18\%  &49.15\%\\\cline{3-13}
    & & D-RTE & 0.89  &1.10  &2.62  &0.29  &0.49  &0.24  &{\bf }0.13  &55.57\%  &73.45\%  &45.48\% \\ \midrule

    \multirow{6}{*}{RoNIN} & \multirow{3}{*}{Seen} & ATE & 7.89  &26.64  &16.82  &5.11  &{\bf 3.99}  &6.18  &4.62  &9.49\%  &-15.81\%  &25.23\% \\\cline{3-13} 
    & & T-RTE & 5.30  &23.82  &19.50  &3.05  &2.83  &3.27  &{\bf 2.81}  &7.70\%  &0.69\%  &13.91\%   \\\cline{3-13}
    & & D-RTE & 0.42  &0.98  &4.99  &0.22  &0.19  &0.20  &{\bf 0.18}  &18.94\%  &2.75\%  &10.15\%   \\\cline{2-13}
    & \multirow{3}{*}{Unseen} & ATE & 7.62  &23.49  &15.75  &8.73  &5.76  &7.49  &{\bf 5.61}  &35.77\%  &2.60\%  &25.11\% \\\cline{3-13}
    & & T-RTE & 5.12  &23.07  &19.13  &4.87  &4.50  &4.70  &{\bf 4.48}  &8.04\%  &0.42\%  &4.61\%      \\\cline{3-13}
    & & D-RTE & 0.43  &1.00  &5.37  &0.29  &{\bf 0.25}  &0.26  &{\bf 0.25}  &12.63\%  &0\%  &4.83\%    \\ \midrule

    \multirow{6}{*}{IDOL} & \multirow{3}{*}{Seen} & ATE & 21.54  &18.44  &9.79  &4.57  &4.44  &4.68  &{\bf 2.90}  &36.49\%  &34.63\%  &37.98\% \\\cline{3-13} 
    & & T-RTE & 14.93  &14.53  &7.97  &1.72  &1.58  &1.77  &{\bf 1.35}  &21.47\%  &14.54\%  &23.46\%\\\cline{3-13}
    & & D-RTE & 1.07  &1.14  &0.97  &0.19  &0.26  &0.18  &{\bf 0.13}  &28.39\%  &48.21\%  &25.12\% \\\cline{2-13}
    & \multirow{3}{*}{Unseen} & ATE & 20.34  &16.83  &9.54  &5.60  &3.81  &5.89  &{\bf 3.69}  &34.19\%  &3.28\%  &37.40\% \\\cline{3-13}
    & & T-RTE & 18.48  &15.67  &9.07  &1.99  &1.67  &2.21  &{\bf 1.65}  &16.73\%  &1.02\%  &25.30\%\\\cline{3-13}
    & & D-RTE & 1.36  &1.31  &1.04  &0.20  &0.22  &0.20  &{\bf 0.15}  &25.36\%  &30.14\%  &25.52\% \\ \midrule
    
    \multirow{3}{*}{CTIN} & \multirow{3}{*}{Seen} & ATE & 5.63  &12.05  &4.88  &2.22  &2.39  &2.02  &{\bf 1.28}  &42.25\%  &46.45\%  &36.68\% \\\cline{3-13}
    & & T-RTE & 5.34  &16.39  &4.21  &2.10  &2.01  &1.73  &{\bf 1.29}  &38.54\%  &35.87\%  &25.55\%\\\cline{3-13}
    & & D-RTE & 0.50  &0.79  &0.18  &0.11  &0.16  &0.11  &{\bf 0.08}  &28.91\%  &50.56\%  &24.61\% \\
            
    \bottomrule
  \end{tabular}
  }%
  \vspace{-2mm}
  \caption{Overall Trajectory Prediction Accuracy. The best result is shown in  bold font. \brcomment{need more explain}}
  \label{tab: overall_model_perf}
  \vspace{-6mm}
\end{table*}

\section{Experiments}
\label{sec: evaluation}


We evaluate CTIN on five datasets against four representative prior research works. CTIN was implemented in Pytorch 1.7.1 \cite{paszke2019pytorch} and trained using Adam optimizer \cite{kingma2015adam}. During training, early stopping with 30 patience \cite{prechelt1998early, wang2020neural} is leveraged to avoid overfitting according to model performance on the validation dataset. To be consistent with the experimental settings of baselines, we conduct both training and testing on NVIDIA RTX 2080Ti GPU.

\subsection{Dataset and Baseline}

{\bf Dataset} As shown in Table \ref{tab: data_desc}, all selected datasets with rich motion contexts (\eg\; handheld, pocket, and leg) are collected by multiple subjects using two devices: one is to collect IMU measurements and the other provides ground truth, like position and orientation. All datasets are split into training, validation, and testing datasets in a ratio of 8:1:1. For testing datasets except in CTIN, there are two sub-sets: one for subjects that are also included in the training and validation sets, the other for unseen subjects. More information about datasets can be found in the supplementary material.

{\bf Baseline} The selected baseline models are listed below:
\begin{itemize}
    \item {\it Strap-down Inertial Navigation System (SINS):} The subject's position can be obtained from double integration of linear accelerations (with earth's gravity subtracted). To this end, we need to rotate the accelerations from the body frame to the navigation frame using device orientations and perform an integral operation on the rotated accelerations twice to get positions \cite{savage1998strapdown}.
    \item {\it Pedestrian Dead Reckoning (PDR):} We leverage an open-source step counting algorithm \cite{murray2018adaptiv} to detect foot-steps and update positions per step along the device heading direction. We assume a stride length of 0.67m/step.
    \item {\it Robust IMU Double Integration (RIDI):} We use the original implementation \cite{yan2018ridi} to train a separate model for each device attachment in RIDI and OxIOD datasets. For the rest of the datasets, we train a unified model for each dataset separately, since attachments during data acquisition in these datasets are mixed. 
    \item {\it Robust Neural Inertial Navigation (RoNIN):} We use the original implementation \cite{herath2020ronin} to evaluate all three RoNIN variants (\ie\;R-LSTM, R-ResNet, and R-TCN) on all datasets.  
\end{itemize}
\begin{figure*}[ht!]
\centering
\begin{subfigure}{.55\textwidth}
  \centering
  \includegraphics[width=\linewidth]{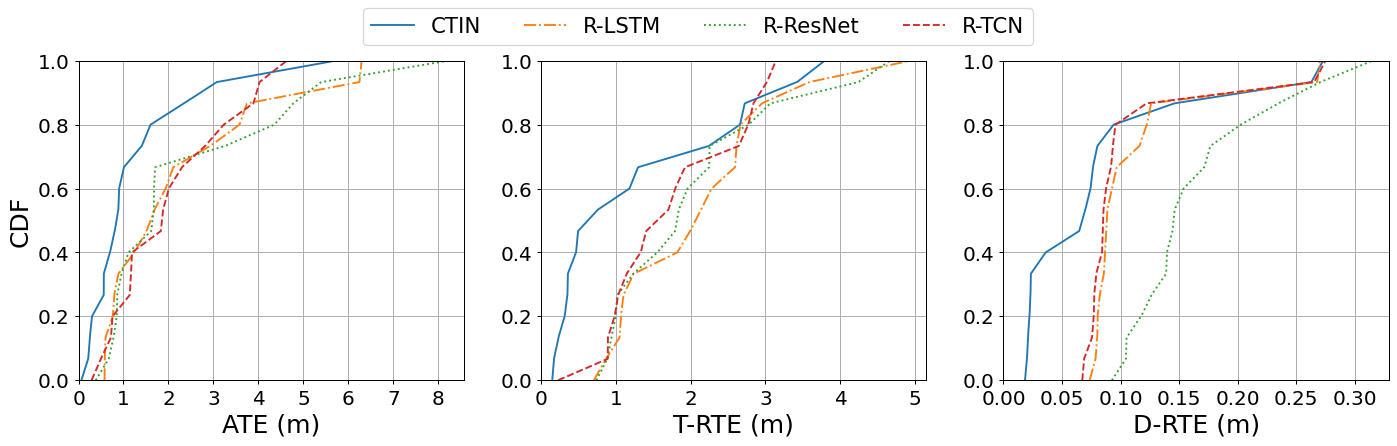}
  \caption{}
  \label{fig:unknot_seen_model_cdf}
\end{subfigure}%
\begin{subfigure}{.45\textwidth}
  \centering
  \includegraphics[width=\linewidth]{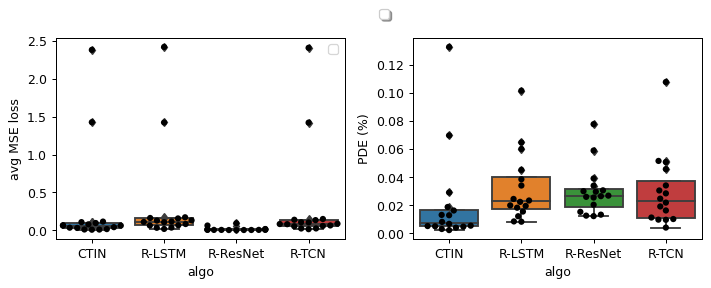}
  \caption{}
  \label{fig:unknot_seen_model_net}
\end{subfigure}
\vspace{-3mm}
\caption{Performance Comparison of CTIN and RoNIN variant models on CTIN dataset}
\label{fig:unknot_perf_overall}
\vspace{-2mm}

\end{figure*}
\begin{figure*}[ht!]
\centering
\begin{subfigure}{.55\textwidth}
  \centering
  \includegraphics[width=\linewidth]{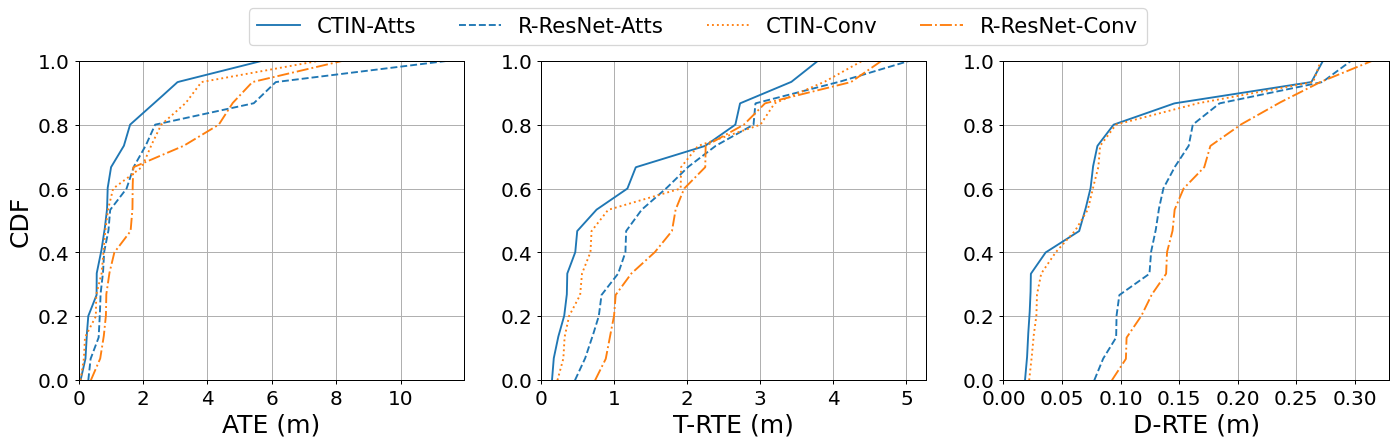}
  \caption{}
  \label{fig:unknot_seen_atts_cdf}
\end{subfigure}%
\begin{subfigure}{.46\textwidth}
  \centering
  \includegraphics[width=\linewidth]{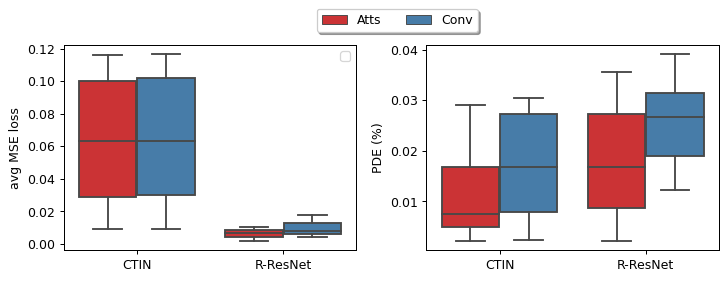}
  \caption{}
  \label{fig:unknot_seen_atts_net}
\end{subfigure}
\vspace{-4mm}
\caption{The effectiveness of proposed attention layers on CTIN dataset. ``*-atts" means CTIN or R-ResNet models with attention functionalities; ``*-Conv" represents the models using a conventional spatial convolution instead.}
\label{fig:unknot_perf_atts}
\vspace{-6mm}
\end{figure*}
\subsection{Evaluation Metrics}
Usually, positions in trajectory can be calculated by performing integration of velocity predicted by CTIN. The major metric used to evaluate the accuracy of positioning is a Root Mean Squared Error (RMSE) with various definitions of estimation error: $RMSE = \sqrt{\frac{1}{m} \sum_{t=1}^m \parallel E_t(x_t, \tilde{x}_t) \parallel}$, where $m$ means the number of data points; $E_t(x_t, \tilde{x}_t)$ represents an estimation error between a position (\ie $x_t$) in the ground truth trajectory at timestamp $t$ and its corresponding one (\ie \;$\tilde{x}_t$) in the predicted path. In this study, we define the following metrics \cite{sturm2011towards}:

\begin{itemize}
    \item \textbf{Absolute Trajectory Error (ATE)} is the RMSE of estimation error: $E_t = x_t - \tilde{x}_t$. The metric shows a global consistency between the trajectories and the error is increasing by the path length.
    \item \textbf{Time-Normalized Relative Traj. Error (T-RTE)} is the RMSE of average errors over a time-interval window span (\ie\;$t_{i}$ = 60 seconds in our case). The estimation error is defined formally as $E_t = (x_{t+t_{i}} - x_t) - (\tilde{x}_{t+t_{i}} - \tilde{x}_t)$. This metric measures the local consistency of estimated and ground truth path. 
    \item \textbf{Distance Normalized Relative Traj. Error (D-RTE)} is the RMSE across all corresponding windows when a subject travels a certain distance $d$, like $d$ is set to 1 meter in our case. The estimation error is given by $E_t = (x_{t+t_d} - x_t) - (\hat{x}_{t+t_d} - \hat{x}_t)$ where $t_d$ is the time interval needed to traverse a distance of $d$.
    \item \textbf{Position Drift Error (PDE)} measures final position (at timestamp $m$) drift over the total distance traveled (\ie\;${\rm traj.\_len}$): $(\parallel x_m - \hat{x}_{m} \parallel) \;/ \;{\rm traj.\_len}$
\end{itemize}
\begin{table}
\centering
\resizebox{1\linewidth}{!}{%
    \begin{tabular}{c|c|c|c|c|c|c}
    \toprule
    \multirow{2}{*}{Model} & \multirow{2}{*}{\shortstack{$\textrm{N}^{\underline{o}}$ of Parameters \\($1 \times 10^6$)}} & \multirow{2}{*}{\shortstack{GFLOP Per Second \\  ($1 \times 10^9$)}} & \multirow{2}{*}{\shortstack{Average GPU time\\ (ms)}} & \multicolumn{3}{c}{Trajectory Error (meter)}\\ \cline{5-7}
    & & & & ATE & T-RTE & D-RTE \\
    \midrule
    CTIN &0.5571& 7.27& 65.96 & {\bf 1.28}& {\bf 1.29}& {\bf 0.08}\\ \hline
    R-LSTM &0.2058& 7.17& 704.23 &  2.22& 2.10& 0.11\\\hline
    R-TCN & 2.0321& 33.17& 19.05 &  2.02& 1.73& 0.11\\\hline
    R-ResNet & 4.6349 & 9.16 & 75.89 &  2.39& 2.01& 0.16 \\
    \bottomrule
    \end{tabular}
  }%
\vspace{-2mm}
\caption{Models' Evaluation Performance on CTIN dataset}
\label{tab:model_perf}
\vspace{-6mm}
\end{table}

\subsection{Overall Performance}

Table \ref{tab: overall_model_perf} shows experimental trajectory errors across entire test datasets. 
It demonstrates that CTIN can achieve the best results on most datasets in terms of ATE, T-RTE, and D-RTE metrics, except for two cases in RoNIN and RIDI datasets. R-TCN can get a smaller T-RTE number than CTIN in the RIDI-unseen test case; R-ResNet reports the smallest ATE of 3.99 for RoNIN-seen. In particular, CTIN improves an average ATE on all seen test datasets by 34.74\%, 21.78\%, and 37.46\%  over R-LSTM, R-ResNet, and R-TCN, respectively; the corresponding numbers for all unseen test datasets are 34.73\%, 3.97\%, and 29.47\%.

The main limitation of RoNIN variants (\ie\;R-LSTM, R-ResNet, and R-TCN) is that they do not capture the spectral correlations across time-series which hampers the performance of the model. Therefore, it is convincing that CTIN achieves better performance over these baselines. Table \ref{tab: overall_model_perf} also shows that CTIN generalizes well to unseen test sets, and outperforms all other models on test sets. PDR shows a persistent ATE due to the consistent and precise updates owing to the jerk computations. This mechanism leads to PDR failure on long trajectories. Over time, the trajectory tends to drift owing to the accumulated heading estimation and the drift would increase dramatically, which results in decentralized motion trajectory shapes. R-LSTM does not show satisfactory results over large-scale trajectories. The margin of the outperforms of CTIN compared to R-LSTM and R-TCN is notable. The results for SINS show a large drift that highlights the noisy sensor measurements from smartphones.

\begin{figure}
    \centering
    \includegraphics[width=\linewidth]{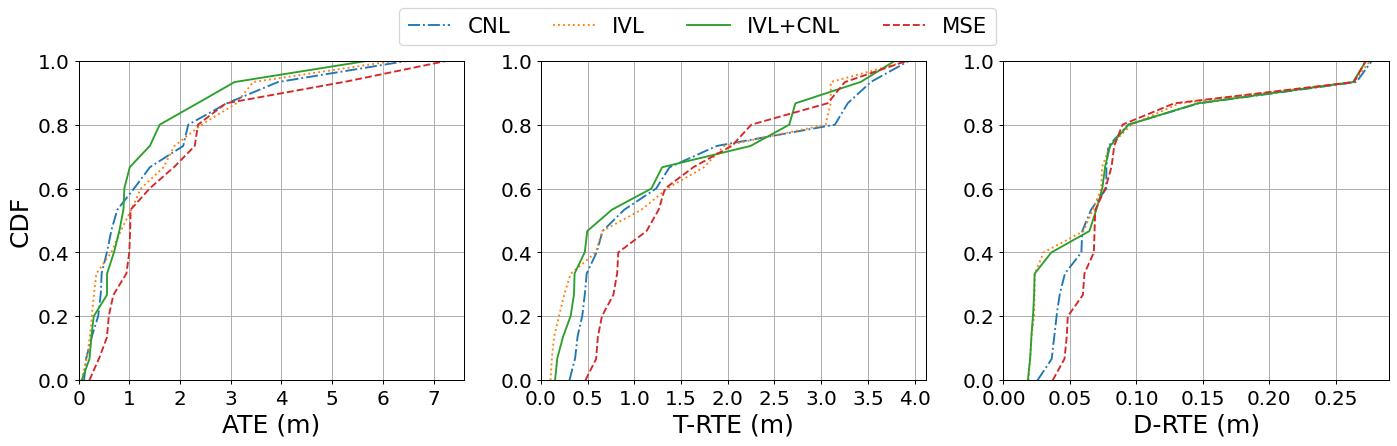}
    \vspace{-4mm}
    \caption{The performance of CTIN network with different loss functions evaluated on CTIN dataset.}
    \label{fig:unknot_perf_loss}
    \vspace{-6mm}
\end{figure}

\subsection{\bf Ablation Study}

In this section, we only evaluate model behaviors, the effectiveness of the attention layer, and loss functions used in CTIN on the CTIN dataset. Please refer to supplementary material for more information about experimental settings and visualizations for the rest of the datasets.

{\bf Model Behaviors.} The experimental results about performance comparisons between CTIN and three RoNIN variants are shown in Figure \ref{fig:unknot_perf_overall} and Table \ref{tab:model_perf}. In Figure \ref{fig:unknot_seen_model_cdf}, each plot shows the cumulative density function (CDF) of the chosen metric on the entire test datasets. The blue line of CTIN is steeper than other plots, which indicates that CTIN shows significantly lower overall errors than all RoNIN variants for all presented metrics. 
As shown in Figure \ref{fig:unknot_seen_model_net}, although CTIN's overall MSE is higher than R-Resnet and smaller than R-LSTM and R-TCN, its position drift error (\ie \; PDE (\%)) is the smallest (\ie \; the best). In Table~\ref{tab:model_perf}, we show the number of parameters for each model, GFLOPs performed by GPU during testing, the average GPU execution time for testing a sequence of IMU samples~(excluding the time to load data and generate trajectories after model prediction) and trajectory errors. Overall, CTIN possesses a significantly smaller number of parameters than R-TCN and R-ResNet, and more parameters than R-LSTM, achieving a competitive runtime performance with lower trajectory errors in a real deployment. Therefore, CTIN performs better than all RoNIN variants. 


{\bf Attention Effectiveness.} In this paper, we propose a novel attention mechanism to exploit local and global dependencies among the spatial feature space, and then leverage the multi-head attention layer to combine spatial and temporal information for better accuracy of velocity prediction. To evaluate their effectiveness, we conduct a group of experiments using CTIN/R-ResNet and their variant without/with the capability of attention mechanism. The experimental results are shown in Figure \ref{fig:unknot_perf_atts}. Figure \ref{fig:unknot_seen_atts_cdf} shows that {\it CTIN-Atts} and {\it R-ResNet-Atts} models outperform the models without attention layer. Furthermore, {\it CTIN-Atts} perform the best for all metrics, and the performance of {\it CTIN-Conv} is better than all R-ResNet variants. In Figure \ref{fig:unknot_seen_atts_net}, {\it CTIN-Atts} and {\it R-ResNet-Atts} have lower average MSE loss of velocity prediction and smallest PDE than {\it CTIN-Conv} and {\it R-ResNet-Conv}. Overall, CTIN and R-ResNet can benefit from the proposed attention mechanism. 

{\bf Loss function.} In this section, we evaluate the performance of multi-task loss (\ie\;IVL+CNL) by performing a group comparison experiments using different loss functions, such as mean square error (MSE), Integral Velocity Loss (IVL) and Covariance NLL Loss (CNL), to train the models. As shown in Figure \ref{fig:unknot_perf_loss}, CTIN with a loss of IVL+CNL achieves the best performance for ATE and D-RTE metrics. 


\section{Related Work}
\label{sec: related_work}

{\bf Conventional Newtonian-based solutions} to inertial navigation can benefit from IMU sensors to approximate positions and orientations \cite{kok2017using}. In a strap-down inertial navigation system (SINS) \cite{savage1998strapdown}, accelerometer measurements are rotated from the body to the navigation frame using a rotation matrix provided by an integration process of gyroscope measurements, then subtracted the earth's gravity. After that, positions can be obtained from double-integrating the corrected accelerometer readings \cite{shen2018closing}. However, the multiple integrations can lead to exponential error propagation. To compensate for this cumulative error, step-based pedestrian dead reckoning (PDR) approaches rely on the prior knowledge of human walking motion to predict trajectories by detecting steps, estimating step length and heading, and updating locations per step \cite{tian2015enhanced}. 


{\bf Data-Driven approach.} Recently, a growing number of research works leverage deep learning techniques to extract information from IMU measurements and achieve competitive results in position estimation \cite{chen2018ionet, chen2018ionet, herath2020ronin, dugne2021understanding}. IoNeT \cite{chen2018ionet} first proposed an LSTM structure to regress relative displacement in 2D polar coordinates and concatenate to obtain the position. In RIDI \cite{yan2018ridi} and RoNIN \cite{herath2020ronin}, IMU measurements are first rotated from the body frame to the navigation from using device orientation. While RIDI regressed a velocity vector from the history of IMU measurements to optimize bias, then performed double integration from the corrected IMU samples to estimate positions. RoNIN regressed 2D velocity from a sequence of IMU sensor measurements directly, and then integrate positions. 


In addition to using networks solely for pose estimates, an end-to-end differentiable Kalman filter framework is proposed in Backprop KF \cite{haarnoja2016backprop}, in which the noise parameters are trained to produce the best state estimate, and do not necessarily best capture the measurement error model since loss function is on the accuracy of the filter outputs. In AI-IMU \cite{brossard2020ai}, state-space models are married to small CNN models to learn a regression model to generate uncertainty covariance noise measurements using MSE loss on the ground truth velocities. TLIO provides a neural model to regress the velocity prediction and uncertainties jointly \cite{liu2020tlio}. The predictions are further applied in the Kalman filter framework as an innovation, where the covariance noise measurement of the Kalman filter is generated by the same deep model. In IDOL \cite{sun2021idol} two separate networks in an end-to-end manner are exploited. The first model is used to predict orientations to circumvent the inaccuracy in the orientation estimations with smartphone APIs. Next, the IMU measurements in the world frame are used to predict the velocities using the second model.

\section{Conclusion and Future Work}
\label{sec: conclusion}

In this paper, we propose CTIN, a novel robust contextual Attention-based model to regress accurate 2D velocity and trajectory from segments of IMU measurements. To this end, we first design a ResNet-based encoder enhanced by local and global self-attention layers to capture spatial contextual information from IMU measurements, which can guide the learning of efficient attention matrix and thus strengthens the capacity of inertial representation. We further fuse these spatial representations with temporal knowledge by leveraging multi-head attention in the Transformer decoder. Finally, multi-task learning using uncertainty is leveraged to improve learning efficiency and prediction accuracy of 2D velocity. Through extensive experiments over a wide range of inertial datasets (\eg\; RoNIN, RIDI, OxIOD, IDOL, and CTIN), CTIN is very robust and outperforms state-of-the-art models. 

The main limitation of CTIN is to use 3D orientation estimation generated by the device~(\eg~Game Vector), which can be inaccurate. In future work, we will extend CTIN with better orientation estimations. Secondly, although the proposed pipeline of CTIN achieves good accuracy on pedestrian inertial observations, the accuracy of CTIN on vehicle IMU data is not desirable due to the errors in the uncertainty of sensory data such as noisy sensory data, inhomogeneous offset values across devices, and variant environments. 
\clearpage

\section*{Acknowledgments}

This material is based upon work supported by the U.S. Army Combat Capabilities Development Command Soldier Center (CCDC SC) Soldier
Effectiveness Directorate (SED) SFC Paul Ray Smith Simulation \& Training Technology Center (STTC) under contract No. W912CG-21-P-0009. Any opinions, findings, and conclusions, or recommendations expressed in this material are those of the author(s) and do not necessarily reflect the views of the CCDC-SC-SED-STTC.

\bibliography{aaai22}

\clearpage
\appendix
\section{Appendix}

We evaluate the performance of CTIN using four public research datasets  (\ie \; RIDI \cite{yan2018ridi}, OxIOD \cite{chen2018oxiod}, RoNIN \cite{herath2020ronin}, and IDOL \cite{sun2021idol}) and the one collected by our own (\ie\; CTIN). In this supplementary document, we provide the details of our data acquisition protocol, data preparation, and extra experimental results and explanations.

\subsection{Data Description and Acquisition}
\label{sec:data_desc}

For these open-source datasets, we developed data loaders following the protocol in the RoNIN project \cite{herath2020ronin} to load and prepare training/testing datasets. To collect the CTIN dataset, we use the two-device framework for IMU and six-degrees-of-freedom ground truth data acquisition.  One device is used to capture IMU data and the other device is used to collect Google ARCore poses (translation and orientation). We use Samsung Galaxy devices in all the sensory experiments. Loop closure measurement is performed before each sensory experiment to ensure high-quality ground truth poses with low drift. An in-house Android application is installed on the devices for IMU data measurements. We use the calibrated IMU data from the device and further remove the offset from acceleration and gyro data through the sensory data in the table test experiment. The IMU data and ARCore data are captured at 200 HZ and 40 HZ, respectively, which leads to spatial and temporal alignment issues. To resolve them, the device system clock is used as the timestamp for sensor events and time synchronization. ARCore data is interpolated at 200 HZ to synchronize the IMU and ARCore devices. For spatial alignment IMU data, ARCore data have to be represented in the same coordinate system. The camera and IMU local coordinate systems are aligned using the rotation matrix estimated by Kalibr toolbox \cite{rehder2016extending}. The data is captured by 5 subjects and it includes various motion activities constitutes from walking and running. For each sequence, a subject moves for 2 to 10 minutes. The IMU device is mounted to the chest by a body harness and the ARCore device is attached to the hand to have a clear line of sight.

\subsection{Data Preparation}


During training, we use a sliding window (N=200) with an overlapping step size (20 for OxIOD, 50 for RIDI, and 10 for the rest of the datasets) on each sequence to prepare input 6D IMU samples, ground truth 2D velocities, and 2D positions. In addition, a random shift is applied to a sliding window to enhance the robustness of the model to the indexing of sliding windows. Since ground truth data are provided in the navigation frame and the network can capture a motion model concerning the gravity-aligned IMU frame, IMU samples in each window are rotated from the IMU body frame to the navigation frame using device orientations at beginning of the window. In this study, the navigation frame is defined that $Z$ axis is aligned with the negation of gravity axis and a coordinate frame augmentation agnostic to the heading in the horizontal frame is applied. This will indirectly provide the gravity information to the network, while augmentation of the sample by rotating around the $Z$ axis in the horizontal plane would remove heading observability as it is theoretically unobservable to the data-driven model and the model should be invariant to rotation around the $Z$ axis.

In this study, we design a component of {\it Rotation Matrix Selector} to choose orientation sources automatically for training, validation, and testing. For the RIDI dataset, we use the orientation estimated from IMU for training, validation, and testing; For the OxIOD dataset, we use ground-truth orientations from Vicon during training/validation, and Eular Angle from the device for testing, because of significant erroneous accuracy of estimated orientations. For the RoNIN dataset, we follow up the same procedures in the RoNIN project to choose orientations for training and testing. That's, estimated orientations are used for testing; during training/validation, estimated orientations are selected if the end-sequence alignment error is below 20 degrees, otherwise, orientations from ground-truth are chosen to minimize noise during training. For IDOL and CTIN datasets, we use orientations from ground truth during training, validation, and testing. In addition to using the uncertainty reduction strategy to train the model, we also increase the robustness of the network against IMU measurements noise and bias by random perturbation of samples, since these perturbations can decrease the sensitivity of the network to input IMU errors. The additive bias perturbations for acceleration and gyroscope data are different. The additive sample bias for acceleration and gyroscope is sampled uniformly from the interval $[-0.2, 0.2]\;m/s^2$ and $[-0.05, 0.05] \; rad/s$ for each sample, respectively. The experimental results demonstrate that CTIN can be more generalized than other baselines to wider use cases or other datasets.



\subsection{Settings and Results}

In this study, we propose a unified model with minor different settings for all datasets. Typically, {\it Spatial Encoder} in CTIN is composed of $Nx=1$ encoder layer; {\it Spatial Decoder} also comprises a stack of $Nx=4$ identical decoder layers. {\it Spatial Embedding} uses a 1D convolutional neural network followed by batch normalization and linear layers to learn spatial representations; {\it Temporal Embedding} adopts a 1-layer bidirectional LSTM model to exploit temporal information, and then adds positional encoding provided by a trainable neural network. For the two MLP-based output branches, a simple linear network followed by a layer normalization can achieve desired performance surprisingly. 

\begin{figure*}[ht!]
\centering
\begin{subfigure}{.55\textwidth}
  \centering
  \includegraphics[width=\linewidth]{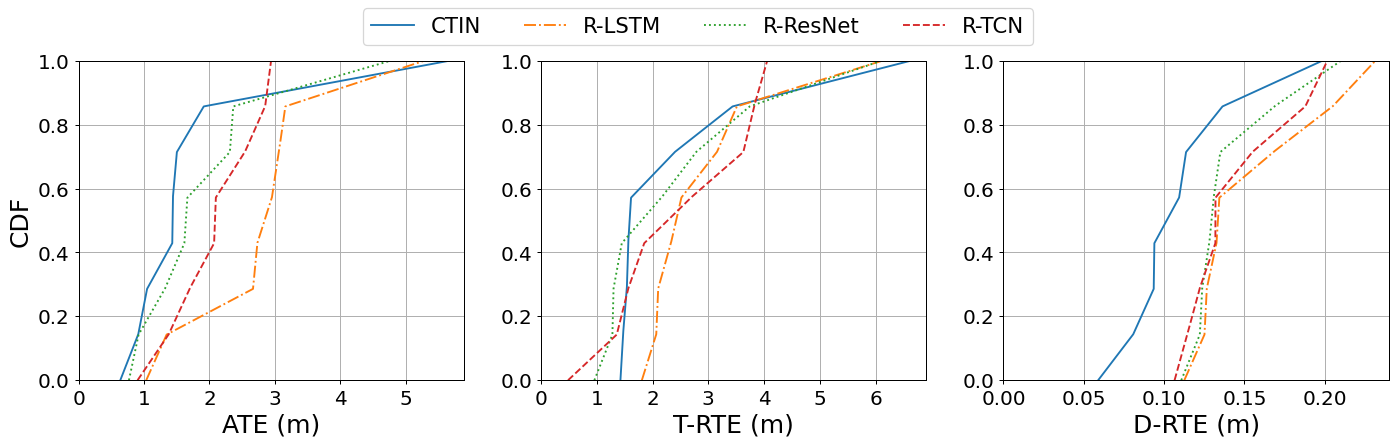}
  \caption{Trajectory Errors on RIDI}
  \label{fig:ridi_unseen_model_cdf}
\end{subfigure}%
\begin{subfigure}{.45\textwidth}
  \centering
  \includegraphics[width=\linewidth]{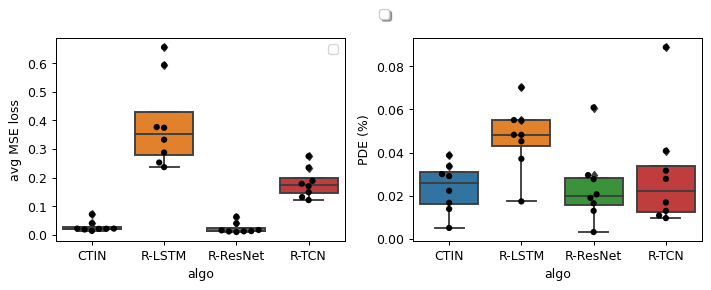}
  \caption{Performance Comparison of Different Models on  RIDI}
  \label{fig:ridi_unseen_model_net}
\end{subfigure}

\hfill

\begin{subfigure}{.55\textwidth}
  \centering
  \includegraphics[width=\linewidth]{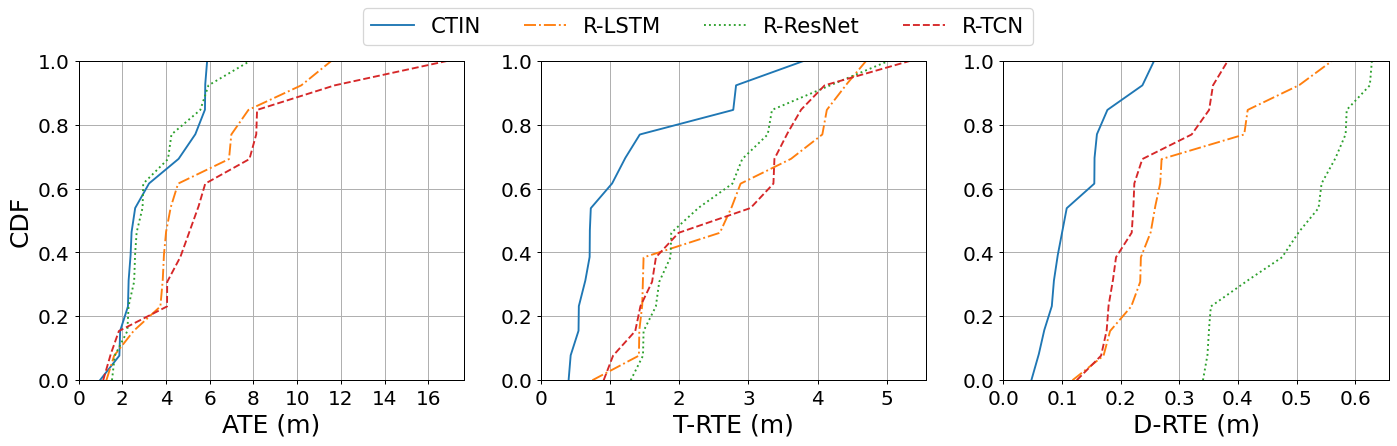}
  \caption{Trajectory Errors on OxIOD}
  \label{fig:oxiod_unseen_model_cdf}
\end{subfigure}%
\begin{subfigure}{.45\textwidth}
  \centering
  \includegraphics[width=\linewidth]{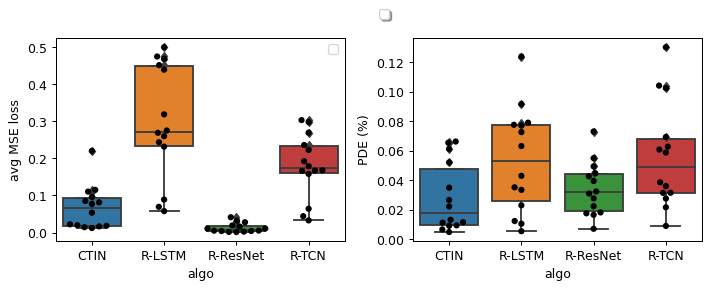}
  \caption{Performance Comparison of Different Models on OxIOD}
  \label{fig:oxiod_unseen_model_net}
\end{subfigure}

\hfill

\begin{subfigure}{.55\textwidth}
  \centering
  \includegraphics[width=\linewidth]{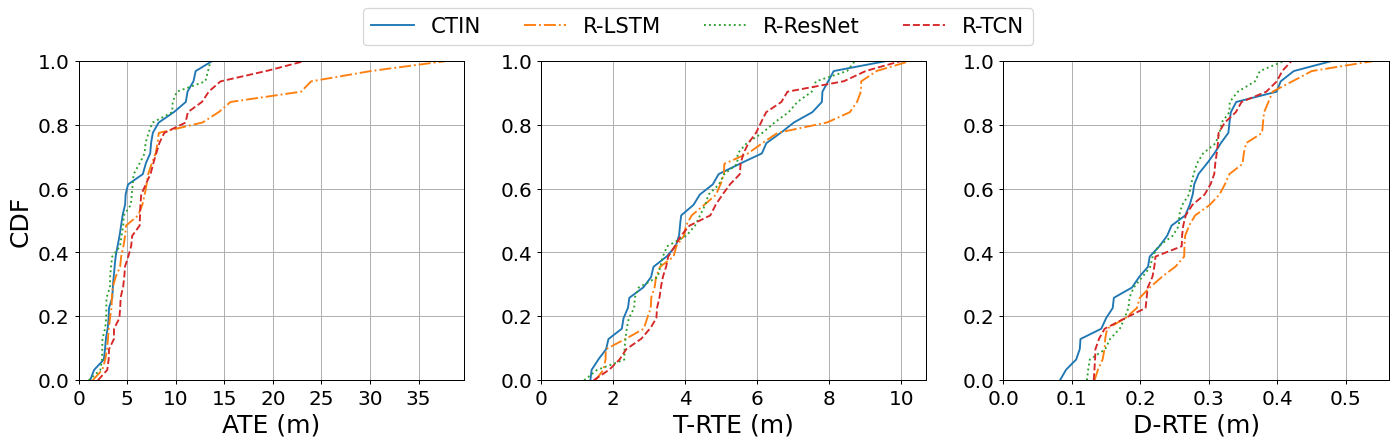}
  \caption{Trajectory Errors on RoNIN}
  \label{fig:ronin_unseen_model_cdf}
\end{subfigure}%
\begin{subfigure}{.45\textwidth}
  \centering
  \includegraphics[width=\linewidth]{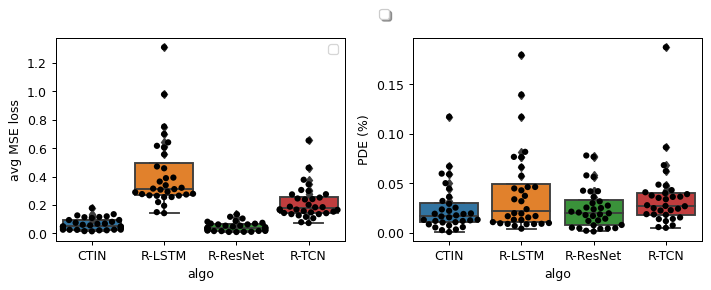}
  \caption{Performance Comparison of Different Models on RoNIN}
  \label{fig:ronin_unseen_model_net}
\end{subfigure}

\begin{subfigure}{.55\textwidth}
  \centering
  \includegraphics[width=\linewidth]{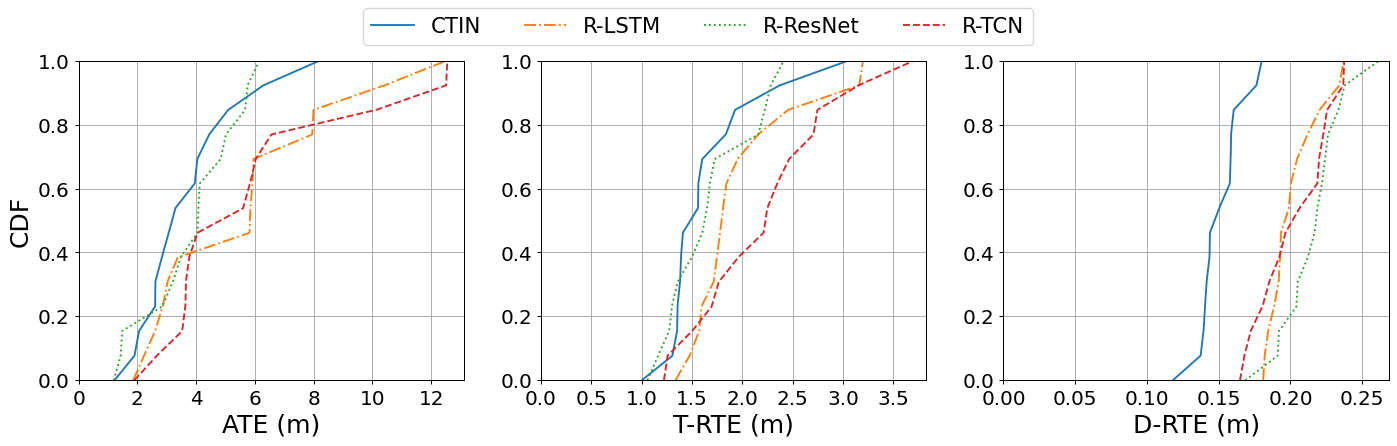}
  \caption{Trajectory Errors on IDOL}
  \label{fig:idol_unseen_model_cdf}
\end{subfigure}%
\begin{subfigure}{.45\textwidth}
  \centering
  \includegraphics[width=\linewidth]{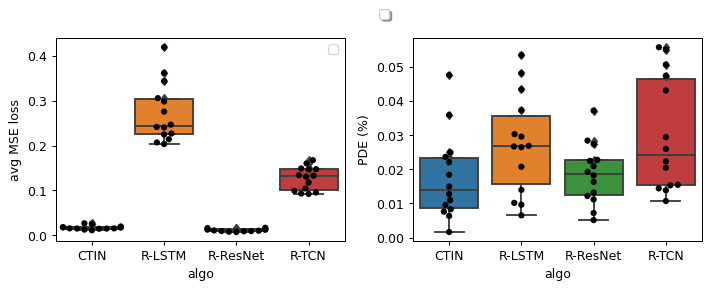}
  \caption{Performance Comparison of Different Models on IDOL}
  \label{fig:idol_unseen_model_net}
\end{subfigure}
\caption{Performance Comparison of CTIN and RoNIN variant models on the selected test dataset.}
\label{fig:all_perf_overall}
\vspace{-3mm}
\end{figure*}

\begin{figure*}[ht!]
\centering
\begin{subfigure}{.55\textwidth}
  \centering
  \includegraphics[width=\linewidth]{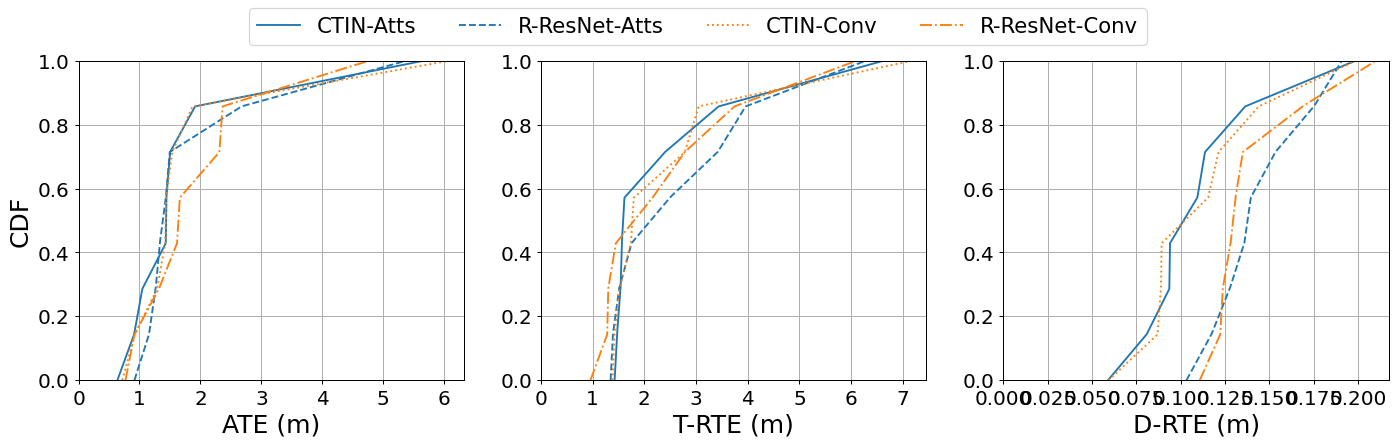}
  \caption{Trajectory Errors on RIDI}
  \label{fig:ridi_unseen_atts_cdf}
\end{subfigure}%
\begin{subfigure}{.46\textwidth}
  \centering
  \includegraphics[width=\linewidth]{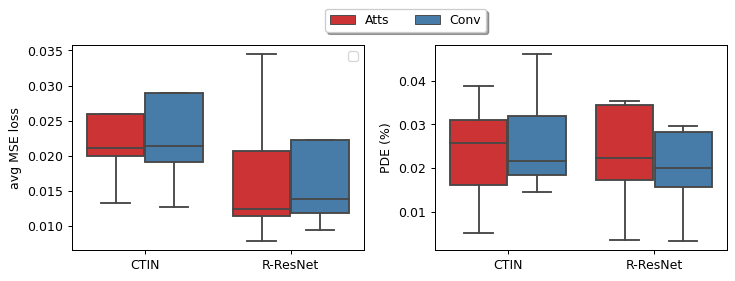}
  \caption{Performance Comparison of Different Models on RIDI}
  \label{fig:ridi_unseen_atts_net}
\end{subfigure}
\begin{subfigure}{.55\textwidth}
  \centering
  \includegraphics[width=\linewidth]{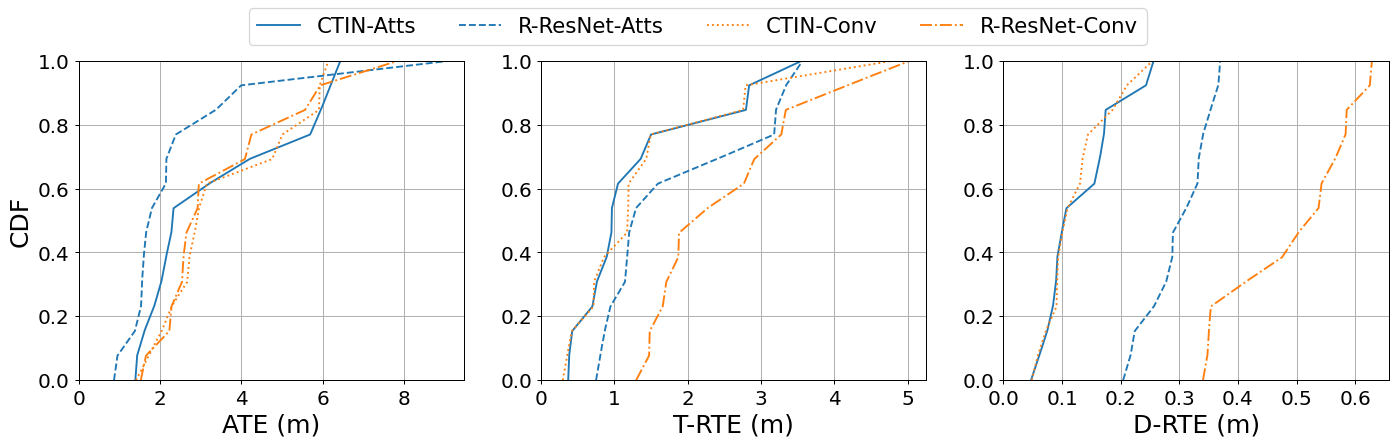}
  \caption{Trajectory Errors on OxIOD}
  \label{fig:oxiod_unseen_atts_cdf}
\end{subfigure}%
\begin{subfigure}{.46\textwidth}
  \centering
  \includegraphics[width=\linewidth]{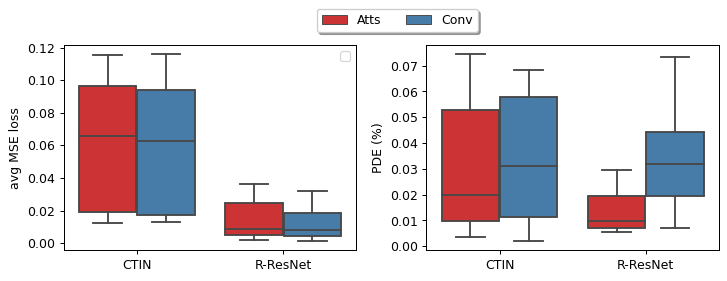}
  \caption{Performance Comparison of Different Models on OxIOD}
  \label{fig:oxiod_unseen_atts_net}
\end{subfigure}
\begin{subfigure}{.55\textwidth}
  \centering
  \includegraphics[width=\linewidth]{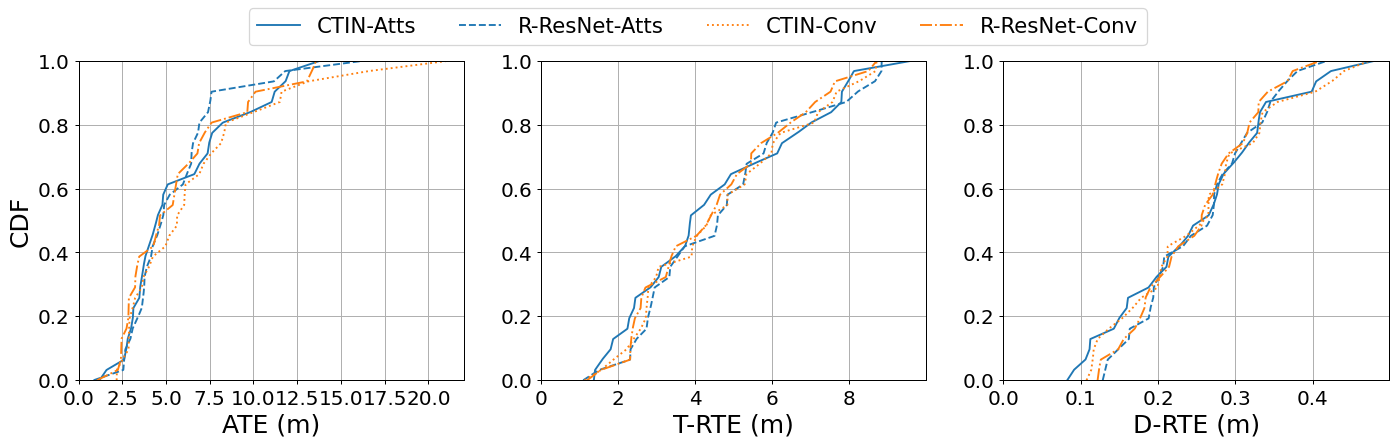}
  \caption{Trajectory Errors on RoNIN}
  \label{fig:ronin_unseen_atts_cdf}
\end{subfigure}%
\begin{subfigure}{.46\textwidth}
  \centering
  \includegraphics[width=\linewidth]{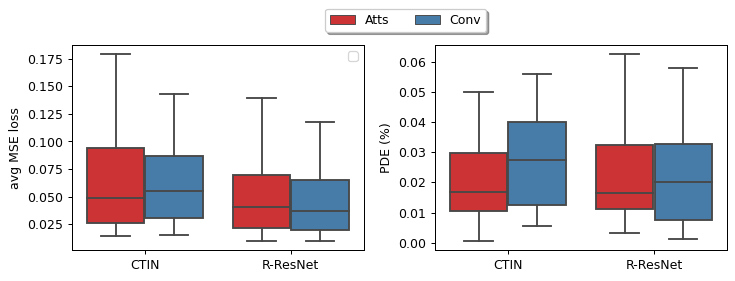}
  \caption{Performance Comparison of Different Models on RoNIN}
  \label{fig:ronin_unseen_atts_net}
\end{subfigure}
\begin{subfigure}{.55\textwidth}
  \centering
  \includegraphics[width=\linewidth]{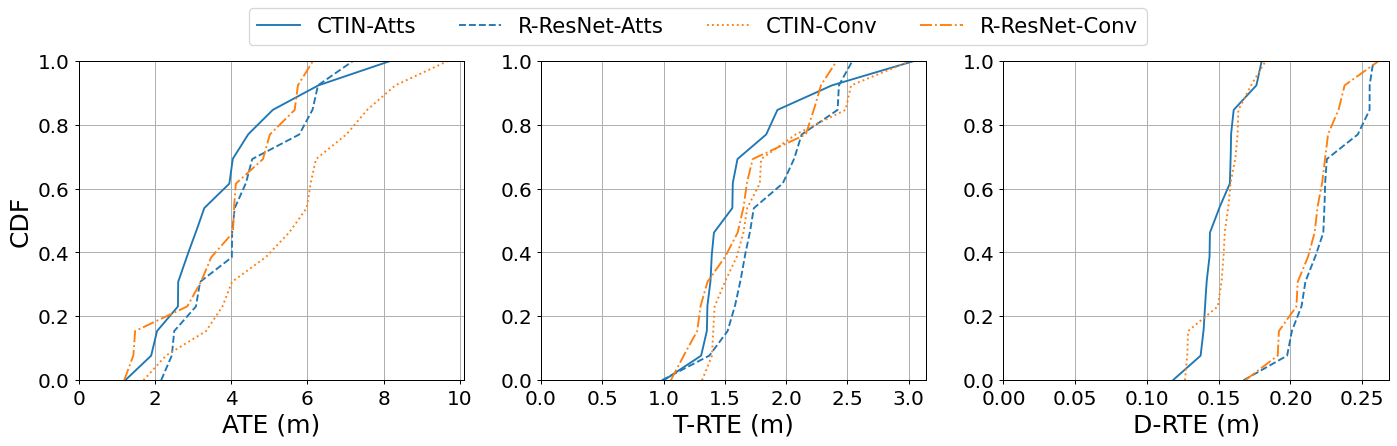}
  \caption{Trajectory Errors on IDOL}
  \label{fig:idol_unseen_atts_cdf}
\end{subfigure}%
\begin{subfigure}{.46\textwidth}
  \centering
  \includegraphics[width=\linewidth]{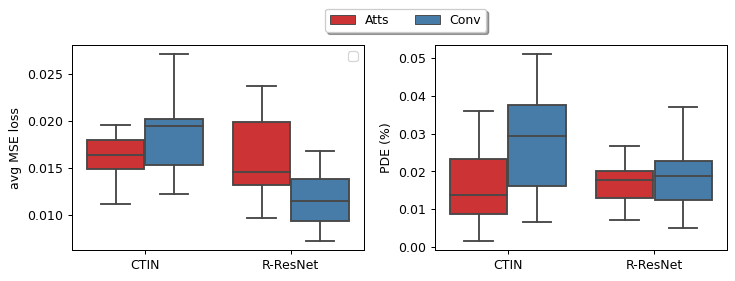}
  \caption{Performance Comparison of Different Models on IDOL}
  \label{fig:idol_unseen_atts_net}
\end{subfigure}
\caption{The effectiveness of proposed attention layers on the selected unseen test dataset. ``*-atts" means CTIN or R-ResNet models with attention functionalities; ``*-Conv" represents the models using a conventional spatial convolution instead.}
\label{fig:all_perf_atts}
\vspace{-3mm}
\end{figure*}

\begin{figure*}[ht!]
\centering
\begin{subfigure}{.5\textwidth}
  \centering
  \includegraphics[width=\linewidth]{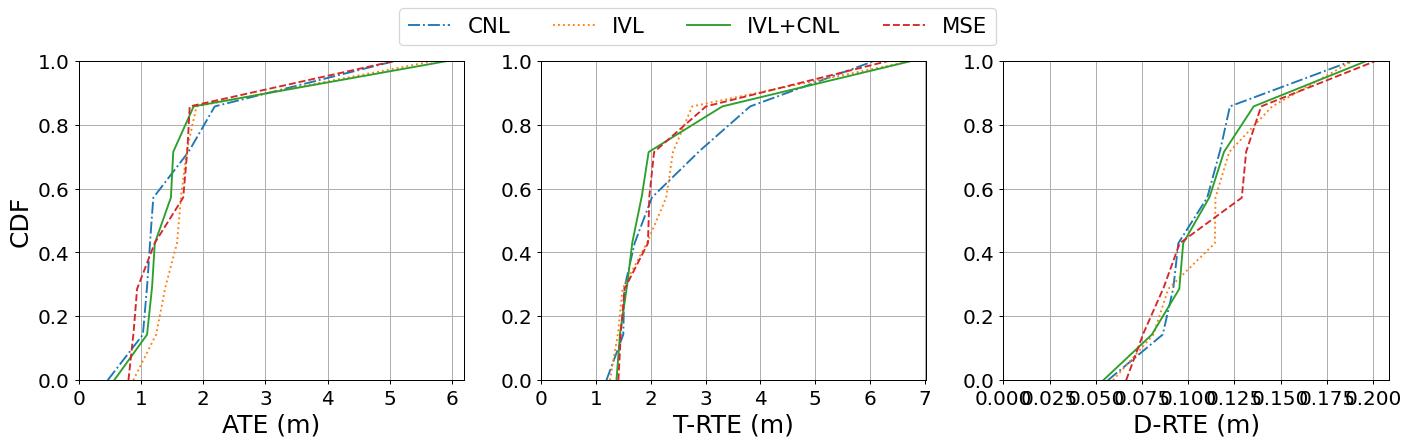}
  \caption{RIDI}
  \label{fig:ridi_unseen_loss_cdf}
\end{subfigure}%
\begin{subfigure}{.5\textwidth}
  \centering
  \includegraphics[width=\linewidth]{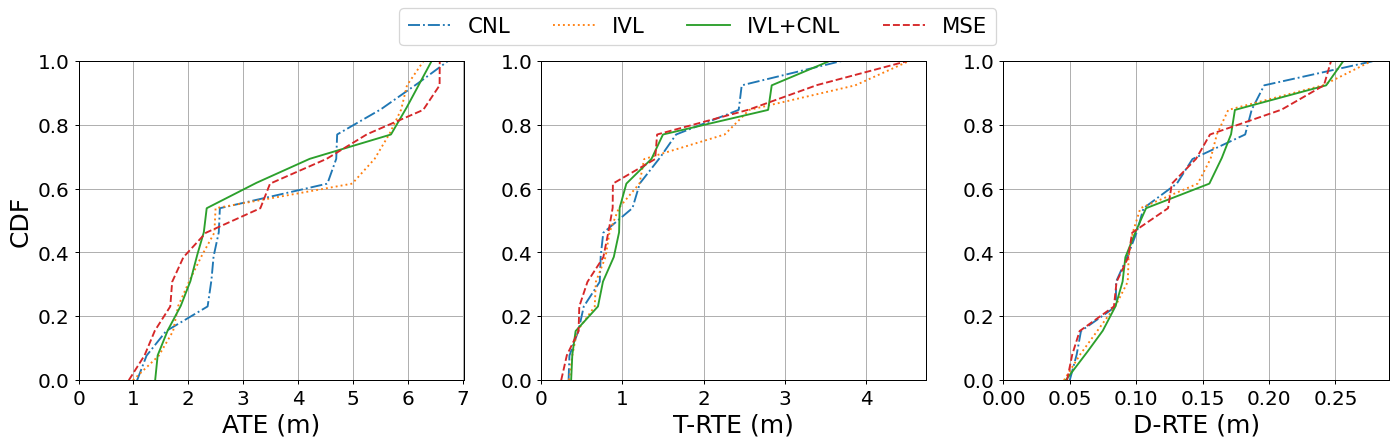}
  \caption{OxIOD}
  \label{fig:oxiod_unseen_loss_net}
\end{subfigure}

\begin{subfigure}{.5\textwidth}
  \centering
  \includegraphics[width=\linewidth]{images/results/loss/ridi_unseen_loss_cdf.png}
  \caption{RoNIN}
  \label{fig:ronin_unseen_loss_cdf}
\end{subfigure}%
\begin{subfigure}{.5\textwidth}
  \centering
  \includegraphics[width=\linewidth]{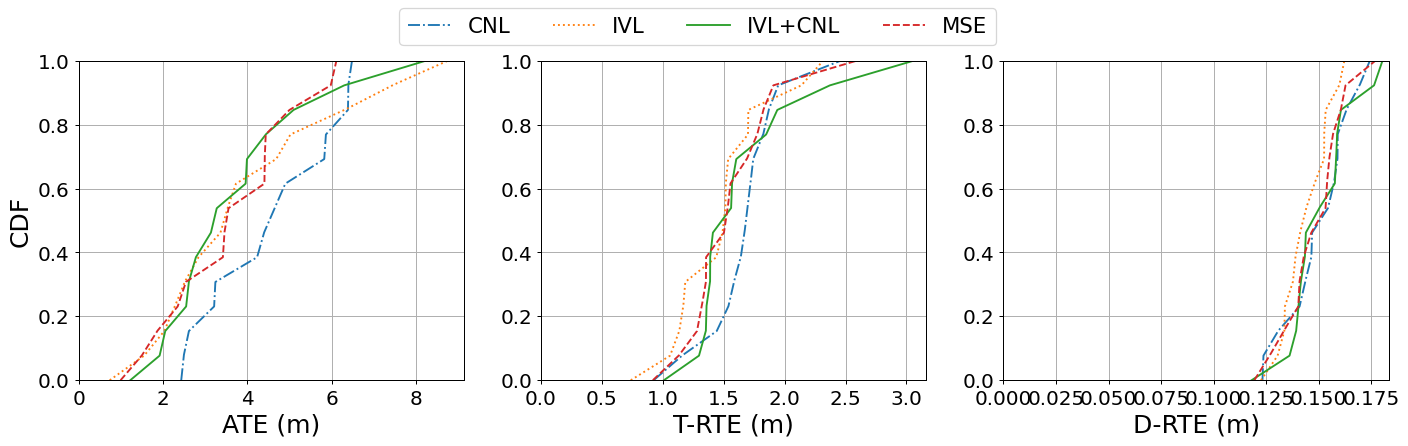}
  \caption{IDOL}
  \label{fig:idol_unseen_loss_net}
\end{subfigure}

\caption{The performance of the CTIN model with different loss functions evaluated on the selected dataset.}
\label{fig:all_perf_loss}
\vspace{-5mm}
\end{figure*}

\begin{figure*}[ht!]
    \centering
    \includegraphics[scale=0.35]{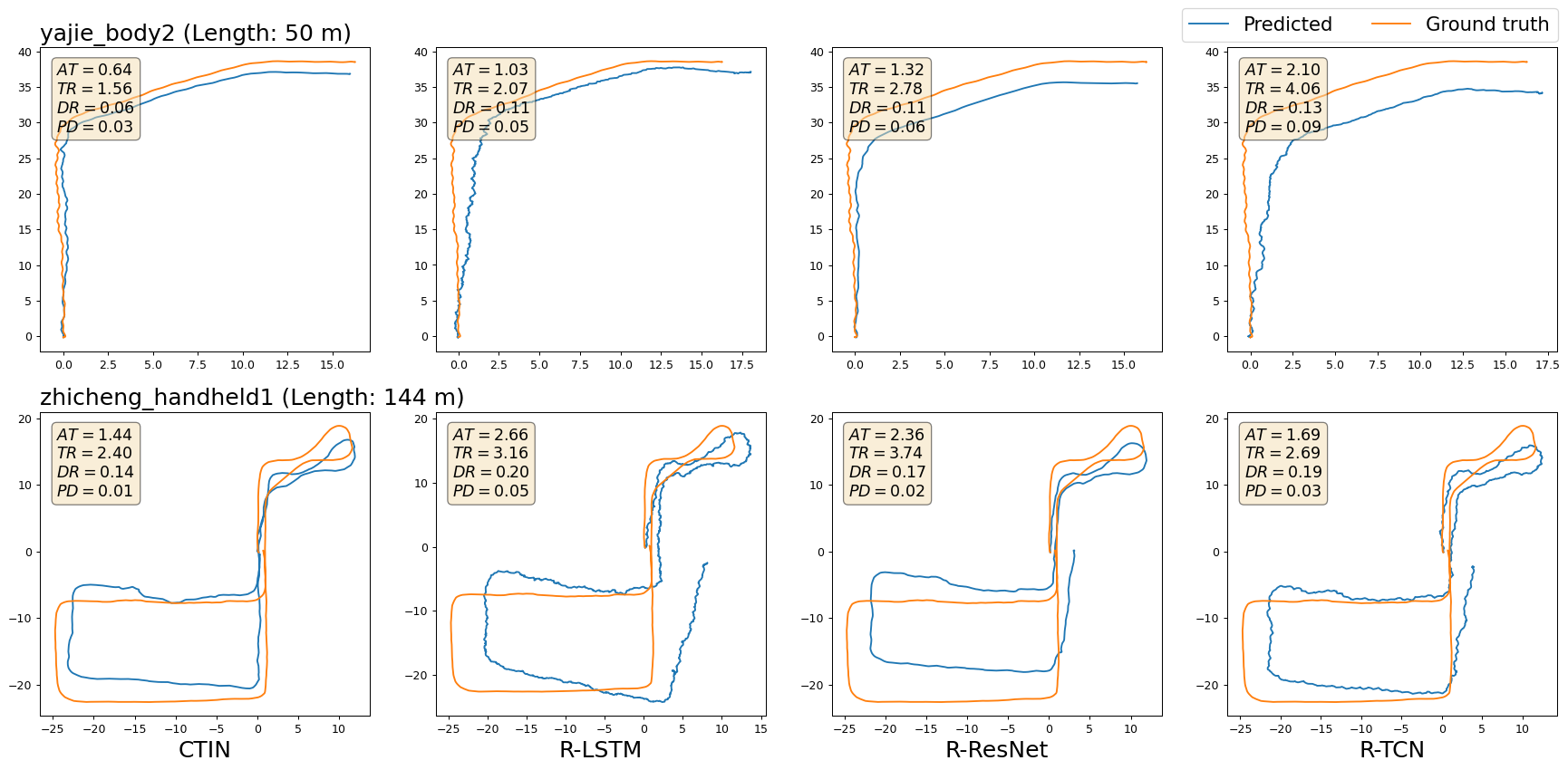}
    \caption{Selected visualizations of trajectories from CTIN and RoNIN variants models on the RIDI dataset. Positional errors are marked within each figure, where ``AT", ``TR", ``DR", and ``PD" denote metrics of ATE, T-RTE, D-RTE, and PDE, respectively. Sub-figures in a row show the visualizations of a selected sequence (named by the title of the first sub-figure) between the ground truth trajectory and predicted ones generated by CTIN, R-LSTM, R-ResNet, and R-TCN, sequentially.}
    \label{fig:ridi_visual}
\end{figure*}

\begin{figure*}[ht!]
    \centering
    \includegraphics[scale=0.35]{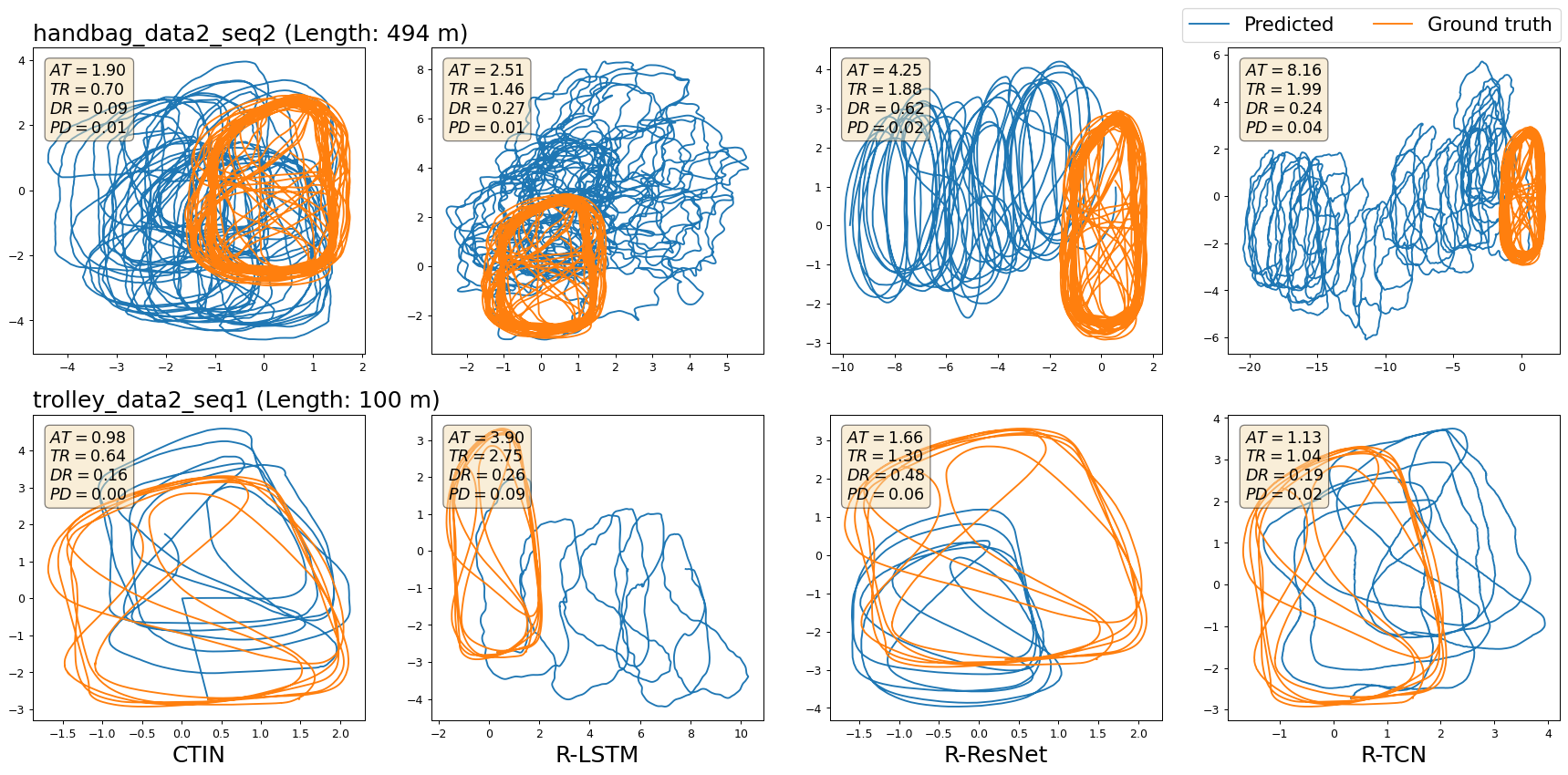}
    \caption{Selected visualizations of trajectories from CTIN and RoNIN variants models on OxIOD dataset. Positional errors are marked within each figure, where ``AT", ``TR", ``DR", and ``PD" denote metrics of ATE, T-RTE, D-RTE, and PDE, respectively. Sub-figures in a row show the visualizations of a selected sequence (named by the title of the first sub-figure) between the ground truth trajectory and predicted ones generated by CTIN, R-LSTM, R-ResNet, and R-TCN, sequentially.}
    \label{fig:oxiod_visual}
\end{figure*}

\begin{figure*}[ht!]
    \centering
    \includegraphics[scale=0.35]{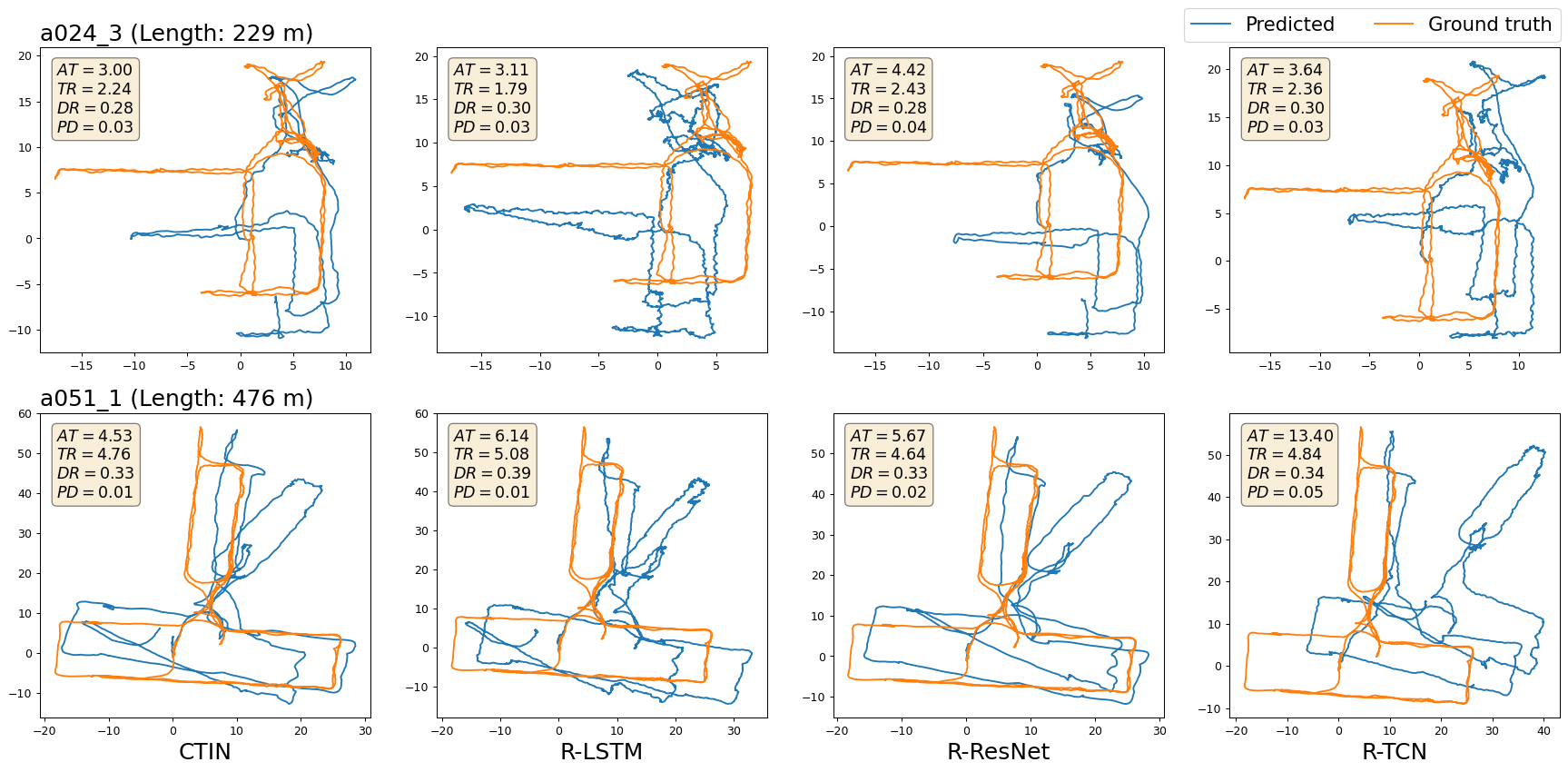}
    \caption{Selected visualizations of trajectories from CTIN and RoNIN variants models on the RoNIN dataset. Positional errors are marked within each figure, where ``AT", ``TR", ``DR", and ``PD" denote metrics of ATE, T-RTE, D-RTE, and PDE, respectively. Sub-figures in a row show the visualizations of a selected sequence (named by the title of the first sub-figure) between the ground truth trajectory and predicted ones generated by CTIN, R-LSTM, R-ResNet, and R-TCN, sequentially.}
    \label{fig:ronin_visual}
\end{figure*}

\begin{figure*}[ht!]
    \centering
    \includegraphics[scale=0.35]{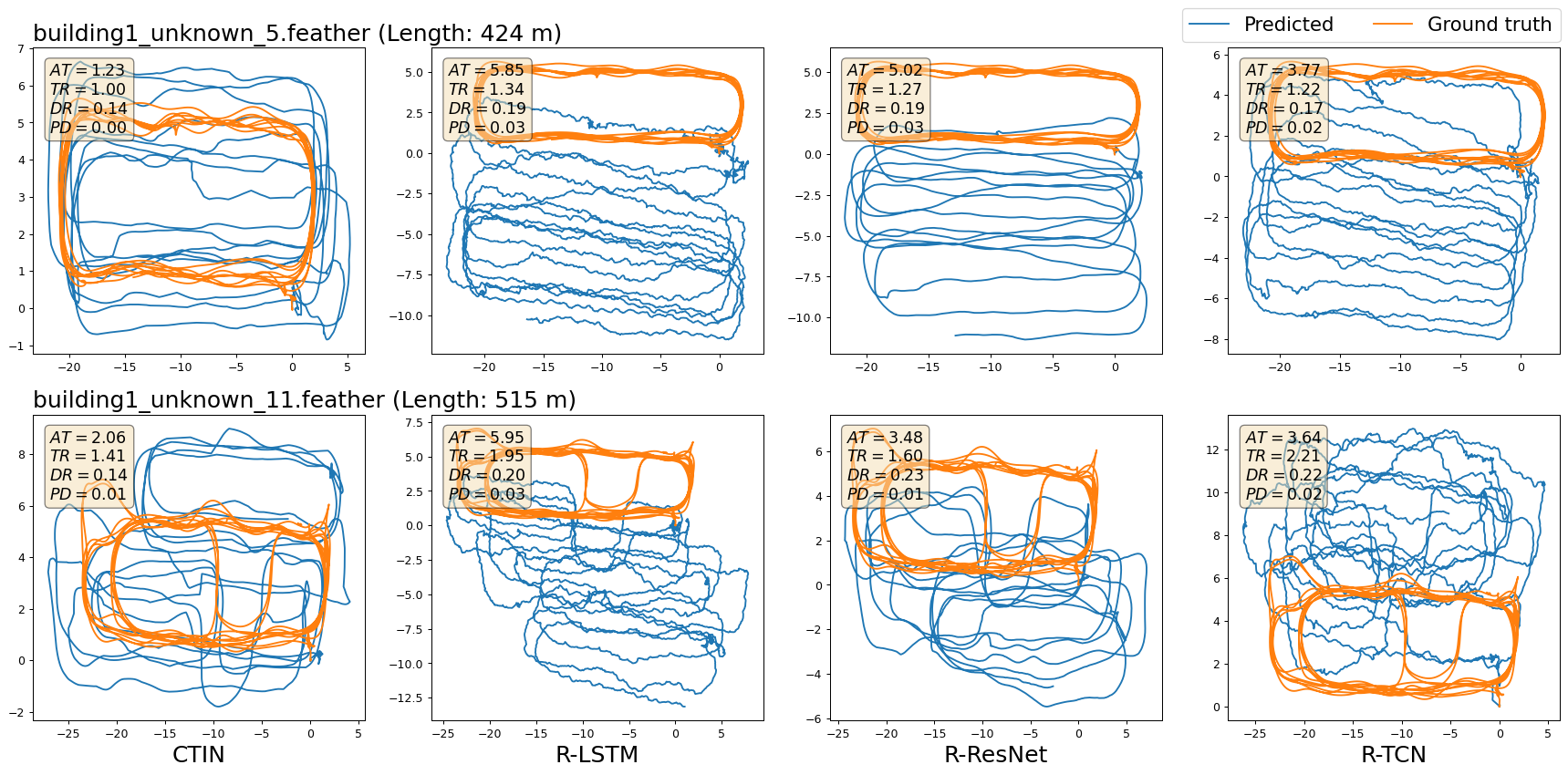}
    \caption{Selected visualizations of trajectories from CTIN and RoNIN variants models on the IDOL dataset. Positional errors are marked within each figure, where ``AT", ``TR", ``DR", and ``PD" denote metrics of ATE, T-RTE, D-RTE, and PDE, respectively. Sub-figures in a row show the visualizations of a selected sequence (named by the title of the first sub-figure) between the ground truth trajectory and predicted ones generated by CTIN, R-LSTM, R-ResNet, and R-TCN, sequentially.}
    \label{fig:idol_visual}
\end{figure*}

CTIN was implemented in Pytorch 1.7.1 \cite{paszke2019pytorch} and trained using Adam optimizer \cite{kingma2015adam} on NVIDIA RTX 2080Ti GPU. During training, we used an initial learning rate of 0.0005, a weight decay value of $1e-6$, and dropouts with a probability of 0.5 for networks in {\it Spatial Encoder} and 0.05 for networks in {\it Temporal Decoder}. Furthermore, early stopping with 30 patience \cite{prechelt1998early} is leveraged to avoid overfitting according to model performance on the validation dataset. The extra experimental results and analysis are listed as follows:

\begin{itemize}
    \item {\bf Overall Performance.} As shown in Figure \ref{fig:all_perf_overall}, CTIN outperforms the three RoNIN variant models (\ie \;R-LSTM, R-ResNet, R-TCN) significantly on RIDI, OxIOD, and IDOL, and lightly better on RoNIN. Specifically, the blue line of CTIN in most sub-figures regarding trajectory errors is steeper than other plots. Sub-figures in the right column of Figure \ref{fig:all_perf_overall}show that CTIN and R-ResNet can obtain lower scores of {\it avg MSE Loss} between ground truth velocity and predicted one, and Position Drift Error ({\it PDE (\%)}), than the other two models. However, the PDE (\%) performance of CTIN is better than R-ResNet, which is consistent with the performance pattern shown in the {\it ATE} metric. For the RoNIN dataset, the best performance is in a tangle of CTIN and R-ResNet. RoNIN is a group of 276 sequences, which is collected by 100 different subjects who perform various motion activities as will. Technically, this dataset should be more comprehensive than others. Unfortunately, only 50\% of the dataset is released by authors, and these 138 (=$276 \times 50\%$) sequences may be gathered by total different 100 subjects, which leads to a significant difference of motion context, and various IMU sensor bias and noise. Therefore, it is difficult for CTIN to learn repeated and shared patterns from this undesired dataset. During training, we also perform random perturbations on the sensor bias, CTIN manifests less sensitivity to these input errors and achieves the desired performance.
    
    \item {\bf Effect of Attention.} Overall, the effectiveness of the proposed attention mechanism has been demonstrated in Figure \ref{fig:all_perf_atts}. For trajectory errors shown in the left column of Figure \ref{fig:all_perf_atts}, CTIN and R-ResNet capability of attention mechanism outperform the ones with spatial convolution layers instead, respectively, especially for OxIOD and IDOL datasets. Attention-based models can achieve lower score of {\it Avg MSE Loss} and {\it PDE (\%)}. 
    
    \item {\bf Effect of Loss function.} We expand the experiments on four extra datasets to evaluate the performance of multi-task loss (\ie\;IVL+CNL) by performing a group comparison experiments using different loss functions, such as mean square error (MSE), Integral Velocity Loss (IVL) and Covariance NLL Loss (CNL), to train the models. Figure \ref{fig:all_perf_loss} verifies the performance of CTIN with loss of IVL+CNL. Accordingly, these four-loss functions can achieve similar performance behaviors. CTIN with loss of IVL+CNL achieves better performance in RIDI OxIOD and IDOL. For RoNIN, the performance of CTIN with CNL is the best, and the model with IVL+CNL is better than the rest of the two loss functions. 
    
    \item {\bf Selected Visualization of Trajectory.} Two selected sequences visualization of reconstructed trajectories against the ground-truth for each dataset is shown from Figure \ref{fig:ridi_visual} to Figure \ref{fig:idol_visual}. We only show CTIN and three RoNIN variants methods. For each sequence, we mark it with sequence name and the trajectory length, also report both ATE, T-RTE D-RTE, and PDE of selected approaches. The trajectory with blue color is generated by the models and the orange one is built from ground truth data. Due to the uncertainty of predicted trajectories, there maybe have different shapes of ground truth trajectory for a sequence. For example in Figure \ref{fig:oxiod_visual}, it looks like the shapes of ground truth trajectory for the sequence ``handbag\_data2\_seq2 (Length: 494m)" are different because of different scales of axes. They are the same and use identical data to draw them.


\end{itemize}
Please refer to the caption in each Figure for more explanations.



\begin{figure*}
    \centering
    \includegraphics[scale=0.35]{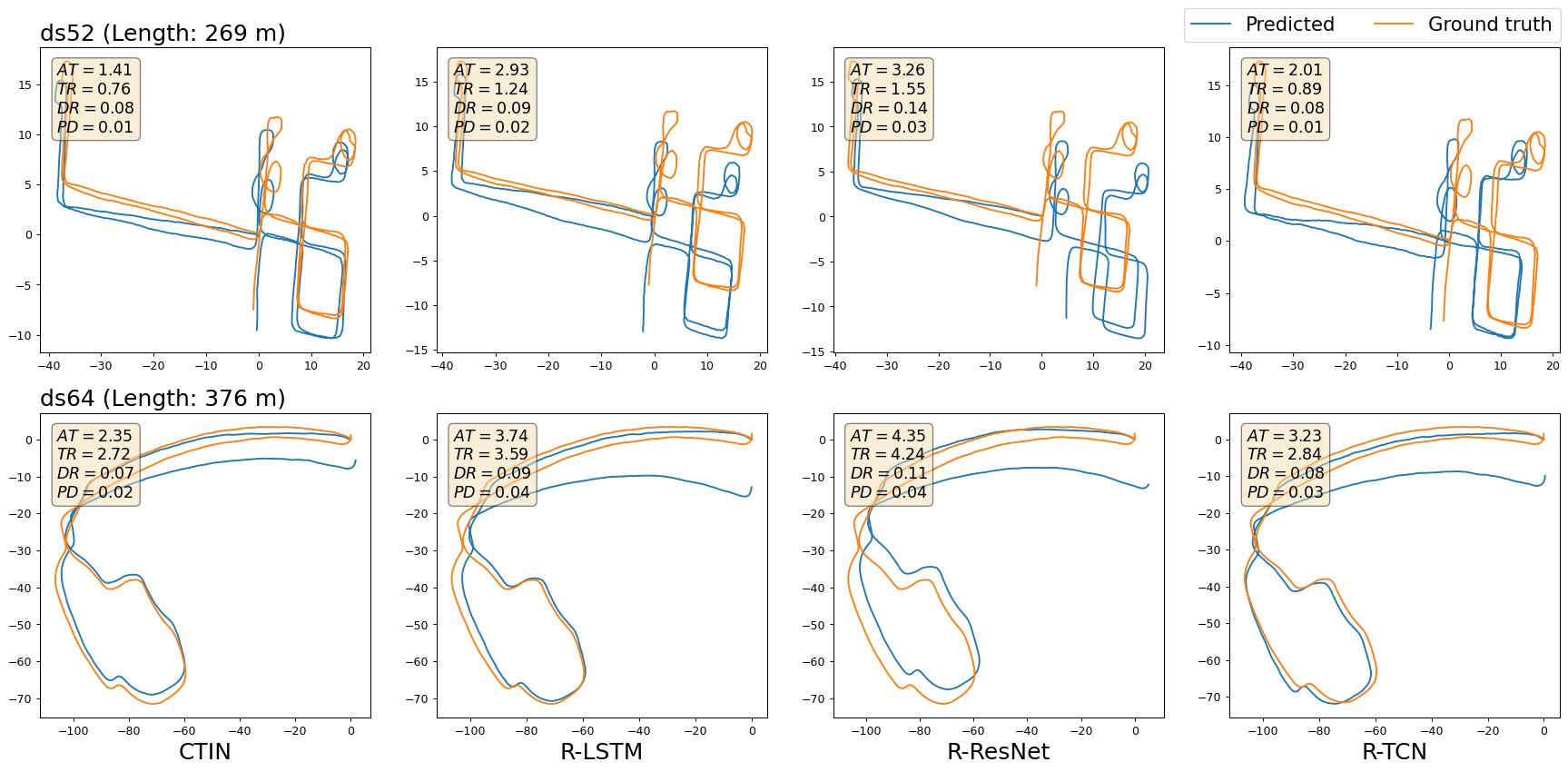}
    \caption{Selected visualizations of trajectories from CTIN and RoNIN variants models on CTIN seen dataset. Positional errors are marked within each figure, where ``AT", ``TR", ``DR", and ``PD" denote metrics of ATE, T-RTE, D-RTE, and PDE, respectively. Sub-figures in a row show the visualizations of a selected sequence (named by the title of the first sub-figure) between the ground truth trajectory and predicted ones generated by CTIN, R-LSTM, R-ResNet, and R-TCN, sequentially.}
    \label{fig:ctin_visual}
\end{figure*}

\clearpage

\end{document}